\documentclass{article}

\usepackage[numbers]{natbib}
\usepackage[preprint]{neurips_2021}

\usepackage[utf8]{inputenc} % allow utf-8 input
\usepackage[T1]{fontenc}    % use 8-bit T1 fonts
\usepackage{hyperref}       % hyperlinks
\usepackage{url}            % simple URL typesetting
\usepackage{booktabs}       % professional-quality tables
\usepackage{amsfonts}       % blackboard math symbols
\usepackage{nicefrac}       % compact symbols for 1/2, etc.
\usepackage{microtype}      % microtypography
\usepackage{xcolor}         % colors
\usepackage{subfigure}
\usepackage{array} 
\usepackage{multirow}
\usepackage{graphicx}
\usepackage{color}

\usepackage{hyperref}
\hypersetup{
	colorlinks=true,
}

\title{Unsupervised Neural Rendering for Image Hazing}

\author{%
  Boyun Li \\
  College of Computer Science\\
  Sichuan University, China\\
  \texttt{liboyun.gm@gmail.com} \\
  \And
  Yijie Lin \\
  College of Computer Science \\
  Sichuan University, China \\
  \texttt{linyijie.gm@gmail.com} \\
  \And
  Xiao Liu \\
  TAL Education \\
  Beijing, China \\
  \texttt{liuxiao15@tal.com} \\
  \And
  Peng Hu \\
  College of Computer Science \\
  Sichuan University, China \\
  \texttt{penghu.ml@gmail.com} \\
  \And
  Jiancheng Lv \\
  College of Computer Science \\
  Sichuan University, China \\
  \texttt{lvjiancheng@scu.edu.cn} \\
  \And
  Xi Peng\thanks{Corresponding author} \\
  College of Computer Science \\
  Sichuan University, China \\
  \texttt{pengx.gm@gmail.com} \\
}

\begin{document}

\maketitle

\begin{abstract}
%  Hazy image generation is an important task in the computer vision. While, existing hazy image generation methods usually rely on expensive professional haze machines or pre-estimated depth maps, and could only generate certain densities and types of haze, which limit applications of it. To overcome above problems, we propose a novel unsupervised method, dubbed as HazeGAN, which is specifically designed for hazy image generation. The major advantages of the proposed method are three-fold. Firstly, it is an unsupervised method that does not use any haze machines or pre-estimated depth information. Secondly, HazeGAN leverages advantages of the atmospheric scattering model, and could generate various density level hazy images from single clean image. Thirdly, the proposed method could generate different types() of haze by introducing hazy images as exemplars. These three advantages enable our method to avoid the expensive data pre-processing, and generate hazy images from any clean images with different densities and types of haze. Extensive experiments show the promising performance of our method comparing with the baseline method in both qualitative and quantitative comparisons. The code will be released on GitHub after the acceptance of the paper. 
Image hazing aims to render a hazy image from a given clean one, which could be applied to a variety of practical applications such as gaming, filming, photographic filtering, and image dehazing. To generate plausible haze, we study two less-touched but challenging problems in hazy image rendering, namely, i) how to estimate the transmission map from a single image without auxiliary information, and ii) how to adaptively learn the airlight from exemplars, \textit{i.e.}, unpaired real hazy images. To this end, we propose a  neural rendering method for image hazing, dubbed as HazeGEN. To be specific, HazeGEN is a knowledge-driven neural network which estimates the transmission map by leveraging a new prior, \textit{i.e.}, there exists the structure similarity (\textit{e.g.}, contour and luminance) between the transmission map and the input clean image. To adaptively learn the airlight, we build a neural module based on another new prior, \textit{i.e.}, the rendered hazy image and the exemplar are similar in the airlight distribution. To the best of our knowledge, this could be the first attempt to deeply rendering hazy images in an unsupervised fashion. Comparing with existing haze generation methods, HazeGEN renders the hazy images in an unsupervised, learnable, and controllable manner, thus avoiding the labor-intensive efforts in paired data collection and the domain-shift issue in haze generation. Extensive experiments show the promising performance of our method comparing with some baselines in both qualitative and quantitative comparisons. The code will be released on GitHub after acceptance.
\end{abstract}

\section{Introduction}

Image rendering is the process of generating a photorealistic image from a 2D or 3D model. As one typical rendering task, hazy image rendering (HIR) aims at synthesizing fog into photographs, which is crucial to many applications including but not limited to gaming, filming, photographic filters, and image dehazing. For example, there are many hazy scenes in the computer games like ``World of Warcraft'' and the films like ``The Lord of the Rings''. Clearly, it is expensive to establish a real hazy scene in making movies even impossible in making games. In addition, as the inverse problem of image dehazing, image rendering could generate a large realistic hazy dataset which is indispensable to the success of deep learning based dehazing methods.

%images captured in outdoor scenes often suffer from poor visibility, reduced contrasts, fainted surfaces, and color shift, due to the presence of haze. As an inverse process of image dehazing, hazy image rendering could generate a large realistic hazy dataset~\cite{Reside} which is beneficial to the training of image dehazing works. 

To render a realistic hazy image, we consider the classical atmospheric scattering model~\cite{Nayar:1999wo}:
\begin{equation}
\label{eq:physical_model}
  I =  t (x) J + (1-t (x )) A,
\end{equation}
where $I$ is the rendered hazy image, and $J$ is the clean (haze-free) scene radiance, $A$ denotes the global atmospheric light, and $t (x)$ is the transmission matrix defined by $t(x) = e^{ - \beta d(x)}$. Here,  $\beta$ is the manually-specified scattering coefficient of the atmosphere and $d(x)$ is the depth between the scene object and the camera. From the model, one could find that the hazy image is sensitive to depth and airlight. In other words, $A$ and $d(x)$ (or equivalently $t(x)$) are the most important two variables. 

The pioneer image rendering methods often resort to the graphics techniques, \textit{e.g.}, geometry~\cite{VRCAI} or binocular vision~\cite{Stereo}. In brief, the methods first reconstruct a 3D scene from a given 2D image and then synthesize the haze into the scene followed by 3D-to-2D flattening. Such a paradigm has suffered from the following limitations. First, they need some auxiliary information to estimate the depth for 3D reconstruction, \textit{e.g.}, stereo pairs~\cite{Stereo} or manual annotations~\cite{VRCAI}. Second, they often determine the airlight in a handcrafted way. As a result, to render different foggy scenes, it is inevitable to manually seek the optimal value for each case, which is a highly consuming and daunting task. Moreover, the user-determined airlight may lead to the domain shift issue, \textit{i.e.}, the synthesized hazy images are inconsistent with the real ones. Third, it is expensive and time-consuming to perform 3D reconstruction. To overcome these limitations, some studies propose utilizing the Kinect sensor or pre-trained monocular depth estimation models to obtain the depth information~\cite{Reside}. The limitations of this type of methods are: i) it is difficult to estimate the desired depth precisely; ii) they need pre-estimated depth maps to guide training, \textit{i.e.}, most of them works in a supervised way and their success largely depends on the accurate pre-estimation of the depth. 
%; and iii) they ignore the connections between the depth and the airlight, and thus might lead to undesirable results. 

\begin{figure*}[!t]
	\begin{center}		\includegraphics[width=1\textwidth]{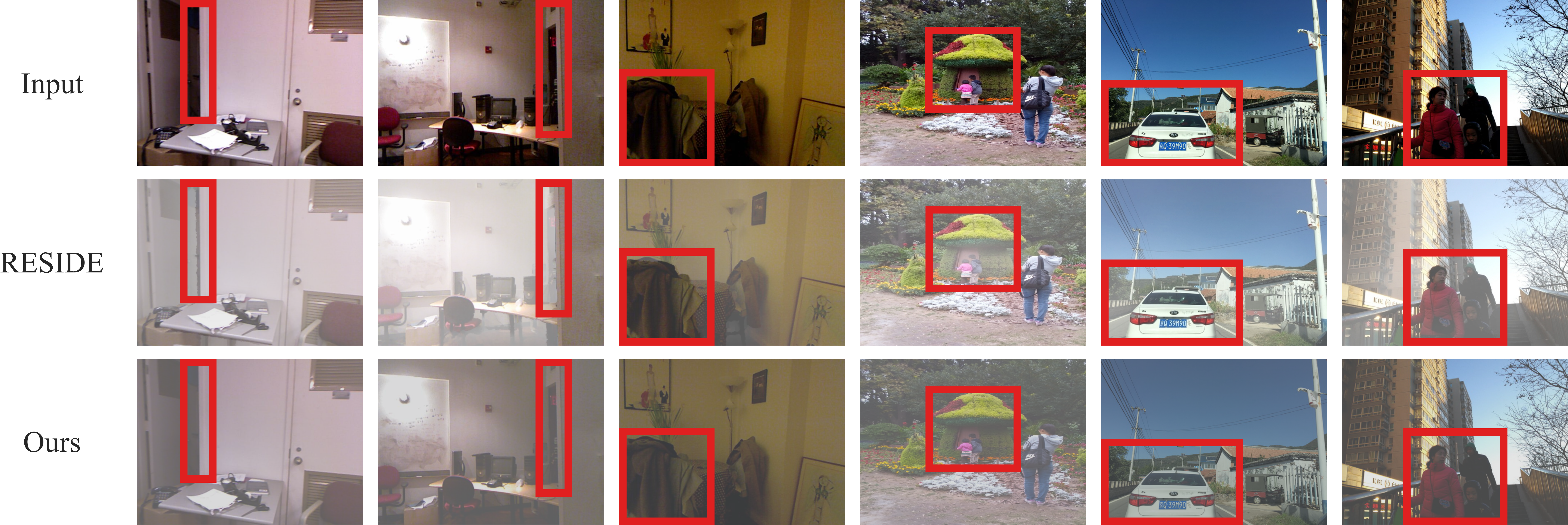}
%		width=\m_width\textwidth, height=\a_height
	\end{center}
	\caption{\label{Figure:1} A visual comparison. To be specific, the first row shows the real-world clean images. The second and third rows are hazy images synthesized by the traditional method~\cite{Reside} and HazeGEN, respectively. From the results, one could find that our method qualitatively performs better than the baseline, \textit{e.g.}, the areas indicated by the red rectangles.}
\end{figure*}

\begin{figure*}[t]
\def \m_wid{0.15}
\def \a_height{1.6cm}
\begin{center}
\subfigure[$x$]{  
\label{fig2a}
\includegraphics[width=\m_wid\columnwidth, height=\a_height]{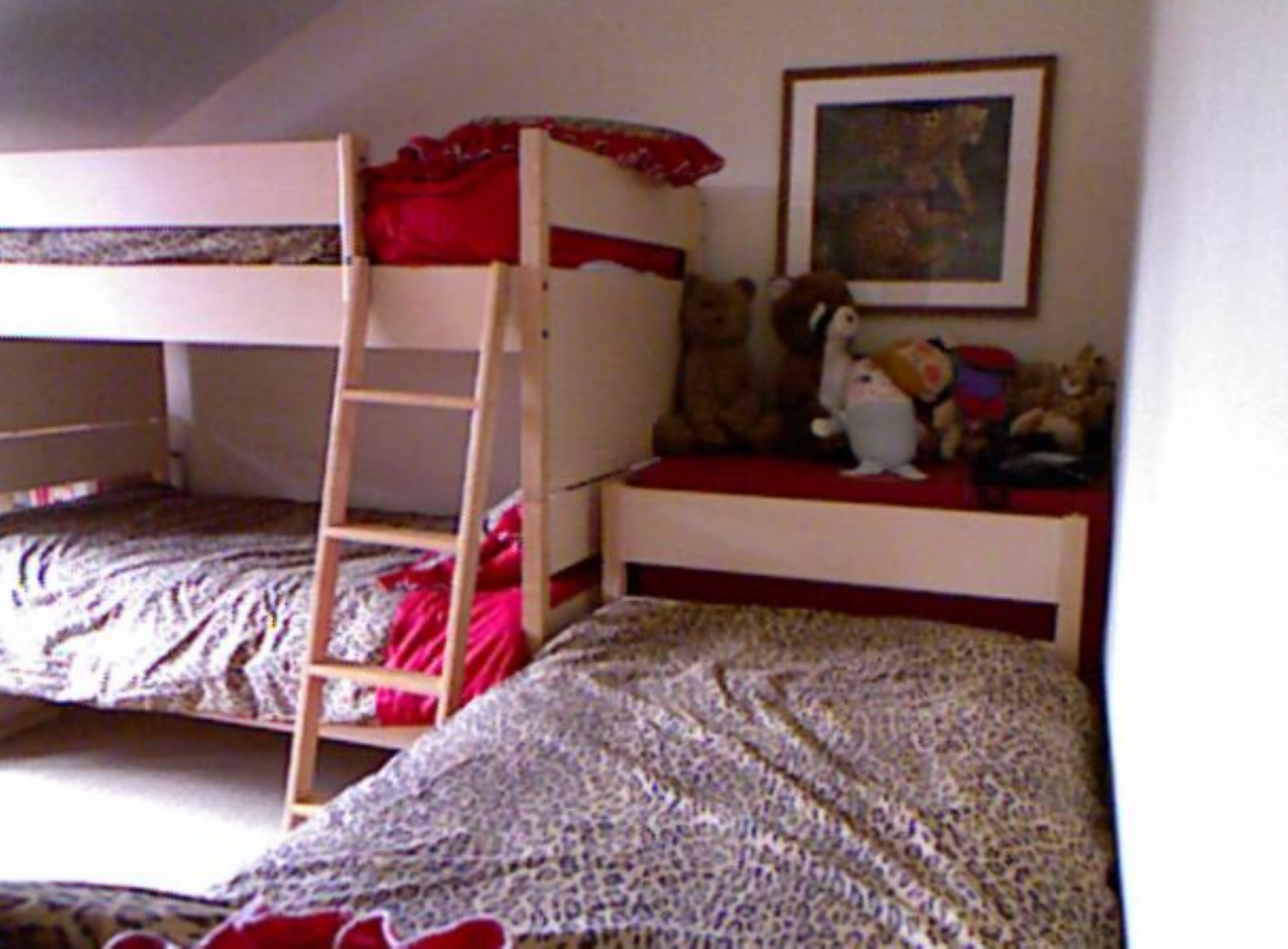}} 
\subfigure[$t(x)$]{
\label{fig2b} 
\includegraphics[width=\m_wid\columnwidth, height=\a_height]{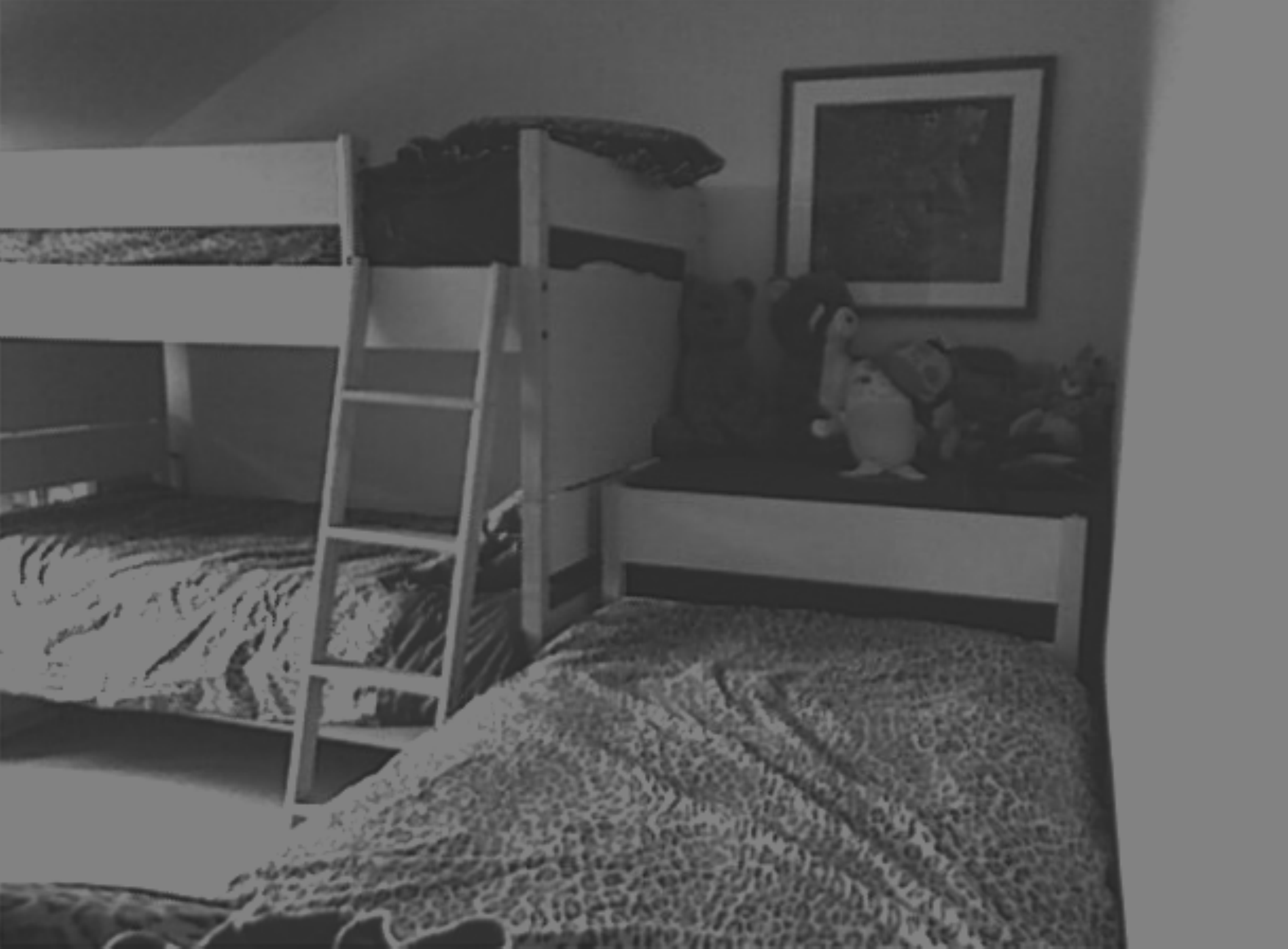}} 
\subfigure[Luminance]{  
\label{fig2c}
\includegraphics[width=\m_wid\columnwidth, height=\a_height]{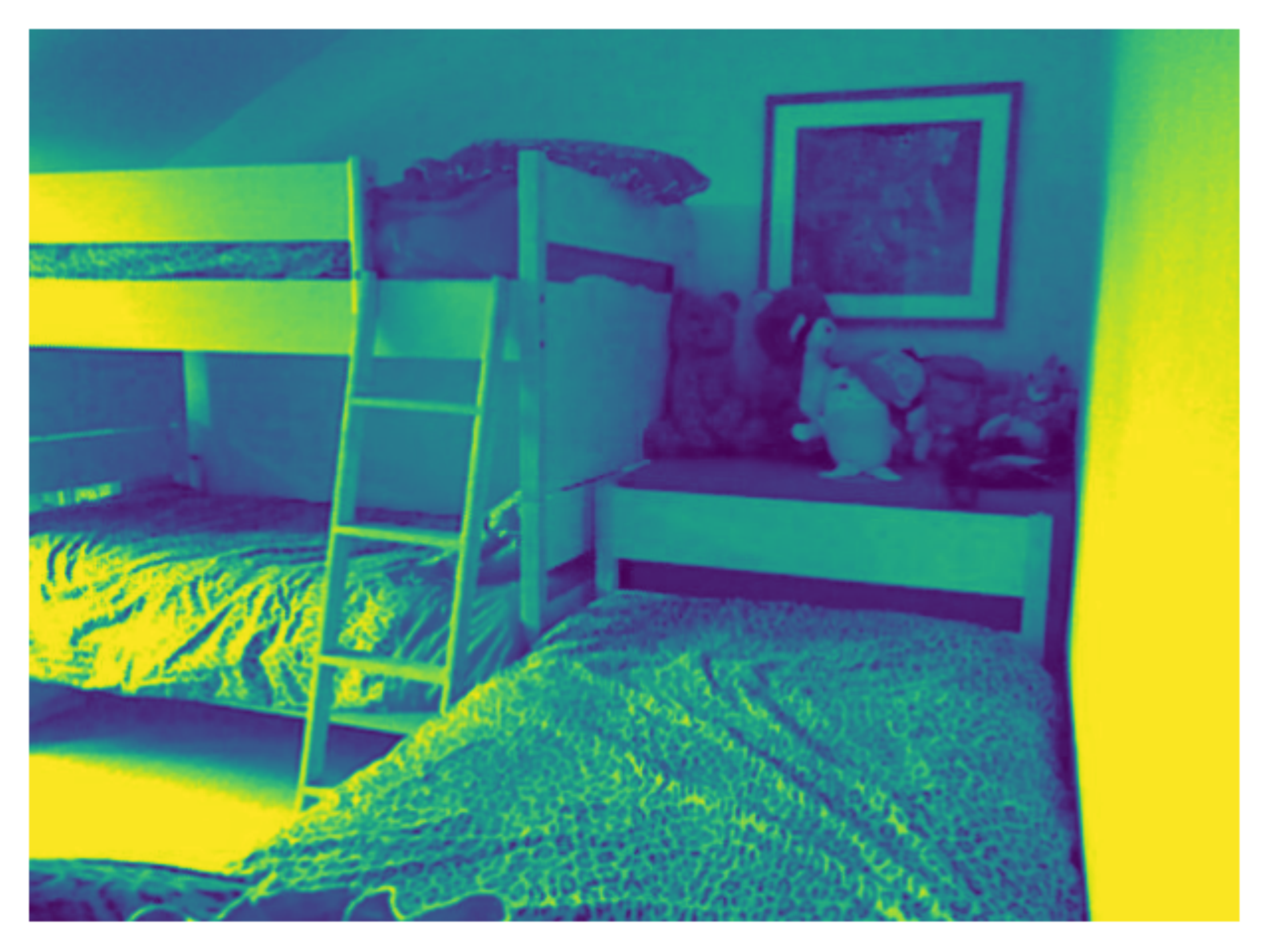}} 
\subfigure[Contour]{  
\label{fig2d}
\includegraphics[width=\m_wid\columnwidth, height=\a_height]{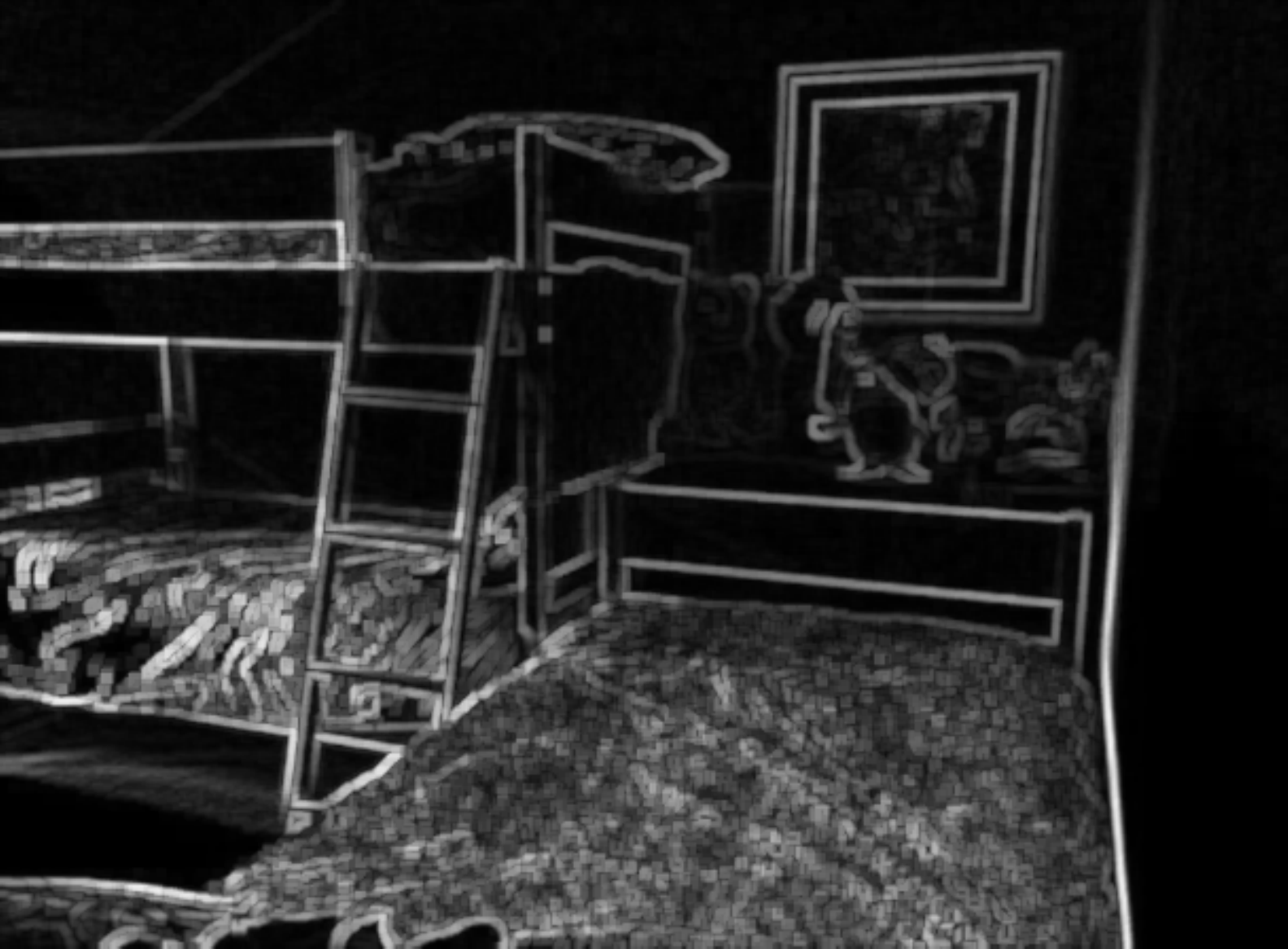}}  
\subfigure[$y$]{  
\label{fig2e}
\includegraphics[width=\m_wid\columnwidth, height=\a_height]{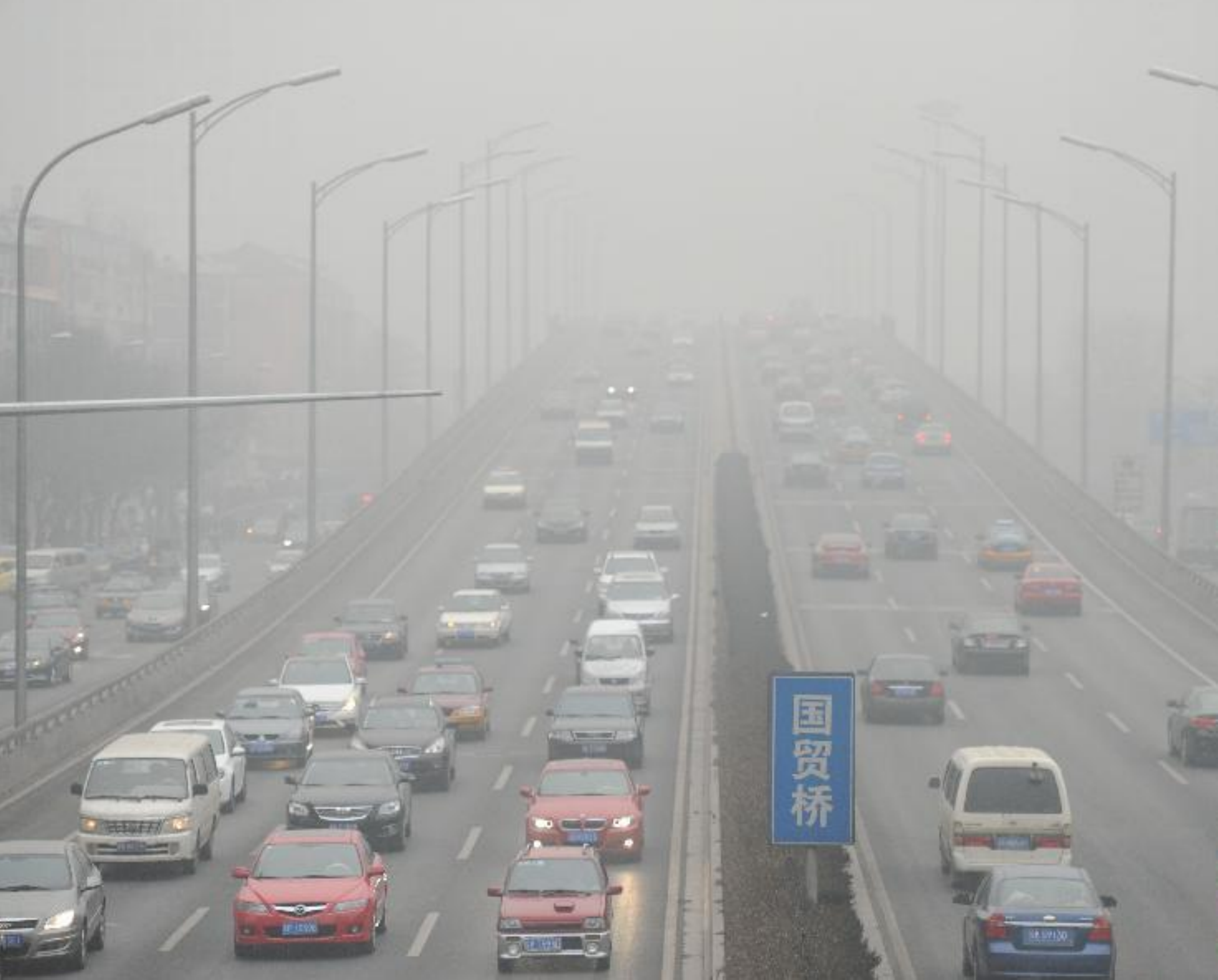}} 
\subfigure[$z$]{  
\label{fig2f}
\includegraphics[width=\m_wid\columnwidth, height=\a_height]{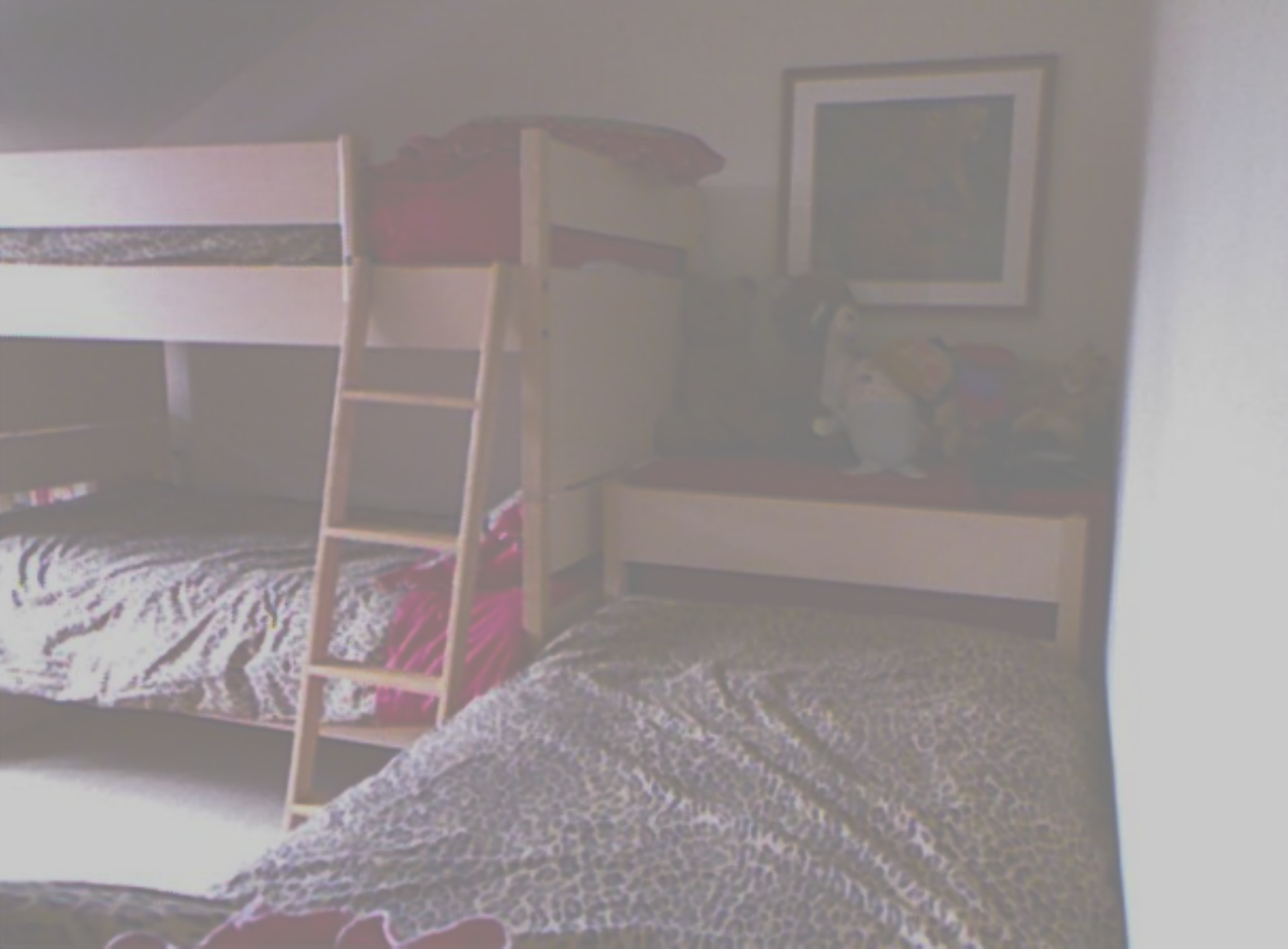}}
\vspace{-0.2cm}
\end{center}
   \caption{Illustrations of our priors. Our method is based on two new priors, \textit{i.e.}, structure similarity prior and airlight consistency prior. The first prior refers to the similarity between the transmission map $t(x)$ (or equivalently depth) and the input image $x$ in terms of luminance and contour. For airlight consistency prior, it refers to the consistency in the airlight distribution  between the rendered hazy image $z$ and the exemplar $y$. Leveraging the above priors, we could obtain $z$ by learning the depth or equivalently the transmission maps from $x$ and the airlight from $y$.
   }
\label{Fig:BasicIdea}
\end{figure*}

In this paper, we aim to solve two less-touched but challenging problems in hazy image rendering. Specifically, i) how to estimate the transmission map $t(x)$ from a single image without auxiliary information, and ii) how to adaptively learn the airlight $A$ from exemplars (unpaired real hazy images) to alleviate the domain-shift issue. 
%in an unsupervised fashion. 
To this end, we propose a novel knowledge-driven neural network for unsupervised haze rendering, dubbed HazeGEN. To be specific, HazeGEN employs a novel prior (\textit{i.e.}, knowledge) to estimate $t(x)$ by leveraging the structure similarity between the depth $d(x)$ (or equivalently $t(x)$) and the input clean image as shown in Fig.~\ref{Fig:BasicIdea}. 
 Thanks to the structure similarity prior, the transmission map estimated by our method could preserve more details. 
To solve the second problem, we design an airlight transfer module which enforces the airlight distribution between the rendered hazy image and the hazy exemplar similar. Thanks to the airlight consistency prior, we could transfer the airlight from a real-world hazy image into the input, thus avoiding labor-intensive parameter selection and the domain-shift issue. 
%To the best of our knowledge, this could be the first work which considers generating the airlight adaptively from exemplars.
%More specifically, HazeGEN proposed a consistency loss and a smoothness loss to keep the structure similarity, and a perceptual loss to keep the airlight invariability. 
The main contributions and novelties of this paper could be summarized as follows:
\begin{itemize}
	\item We propose a novel hazy image rendering method which could be one of the first neural rendering method for image hazing. Our method enjoys the following advantages. First, it is an unsupervised method which does not rely on the   paired data for training. Namely, it estimates the transmission map and airlight in an unsupervised manner. Second, it is a monocular method which only uses the given single image to estimate the transmission map. Third, it is a controllable method which could generate the hazy image with different haze densities and airlights. 
	\item The proposed HazeGEN is a knowledge-driven neural network by using two new priors to build two neural modules. The first prior refers to the structure similarity between the transmission map and the input clean image. The second one refers to the consistency in the airlight distribution between the rendered hazy image and the exemplar.
	\item HazeGEN could be helpful to image dehazing by creating huge training data. Comparing with the widely-used handcrafted data generators, the hazy images rendered by HazeGEN are more consistent with the real situation.
\end{itemize}

\section{Related Work}
This work is close to single image dehazing and hazy image rendering which will be briefly introduced in this section.

\subsection{Single Image Dehazing}

Single image dehazing aims to recover a clean image from an observed hazy image, which has been widely investigated in security cameras, autopilot, and so on. As single image dehazing involves an ill-pose problem, most of the works are based on the atmospheric scattering model (Eq.~\ref{eq:physical_model}). More specifically, a variety of approaches, which have been proposed to estimate the pixel-wise transmission map and global airlight information of the physical model, could be roughly divided into prior- and learning-based methods. Specifically, prior-based methods~\cite{Tan, DCP, CAP} usually approximately estimate the transmission map and airlight by adopting some hand-craft priors such as texture and contrast. 
For example, He et al.~\cite{DCP} found the existence of the dark channel in the outdoor haze-free images and accordingly proposed a dark channel prior to estimating the transmission map and airlight. 
Zhu et al.~\cite{CAP} assumed that the image depth is positively correlated tox the difference between brightness and  saturation, and leveraged this assumption to estimate the transmission map and airlight.

With the development of deep learning, the focus of the community has shifted to building the mapping of a given paired sample~\cite{DehazeNet, MSCNN, AOD-Net, ZID, YOLY, GFN, FFA, DCPDN, zuo1, risheng1}.  More specifically, Cai et al.~\cite{DehazeNet} propose first estimating the transmission map of the input hazy image, and then removing the haze using the physical model. Li et al.~\cite{AOD-Net} directly recover the clean image from the hazy one by reformulating the physical model. It could be seen that the success of these methods heavily relies on large scale hazy-clean data pairs. However, it is difficult even impossible to collect a large scale database with the desirable ground truth due to the changes in the factors such as illumination. Therefore, most of the works resort to collect some clean images and then manually generate the corresponding hazy images with the help of the physical model. 

Different from the above image dehazing works, this paper focuses on generating a hazy image from a clean one rather than in turn. It should be pointed out that image rendering is remarkably different from image dehazing even though their key is the estimation of transmission map and airlight. To be exact, as the airlight could be only estimated from the hazy images, it is difficult to seek the optimal airlight for haze rendering whose input is the clean image.
Besides the numerous applications in the entertainment industries, our HazeGEN could be helpful to improve the dehazing performance because a large number of haze-clean pairs could be synthesized in a data-driven way. Thanks to the learnable and controllable rendering manner, the data generated by HazeGEN could be expected consistent with the real hazy images, thus alleviating even avoiding the domain shift issue.

\subsection{Hazy Image Rendering}

Haze is a typical atmospheric phenomenon caused by the accumulation of the dust, smoke, and other particles. More specifically, these particles will scatter the light ray reflected from the observed scenes objects while introducing the environmental light, thus decreasing scene visibility. Nayar et al.~\cite{Nayar:1999wo}  mathematically analyze and model the haze generation process for the first time, \textit{i.e.}, the well-known atmospheric scattering physical model. 
%Most existing hazy image rendering methods simulate the fog effect under it. As discussed before, to conduct haze rendering, one of the most challenging thing is how to estimate the depth map. 

Based on the physical model, there are some haze synthesis methods have been proposed. The difference and novelty of these methods mainly lie on the depth estimation and the success of them heavily relies on the used auxiliary information, such as stereo pairs or manually annotation.
For instance, \cite{Stereo} estimates the depth using binocular stereo vision. More specifically, they conduct stereo matching on the given stereo pairs to get the disparities which are further transformed into the depth using camera-intrinsic or camera-extrinsic parameters. Although \cite{VRCAI} is independent of the stereo pairs, it requires  manually annotating the objects and the light sources to get the depth map and airlight, respectively.
RESIDE~\cite{Reside} is one of the most popular data sets in recent, which consists of the hazy images generated by the following steps. First, it estimates the depth by using the devices like Kinect or a pretrained depth estimation model. Then, it utilizes the physical model to synthesize the hazy images by randomly generating some scattering and the airlight coefficients. Clearly, the above methods have suffered from the following limitations. More specifically, almost all these methods exhaustively select optimal airlight by humans. As discussed above, such a handcrafted strategy will lead to the domain shift issue and high cost in labor. 
%Second, the data quality heavily depends on the results of the used depth estimation method. Although depth estimation is crucial to hazy image rendering, it is not the only key factor as discussed above. To generate the favorable hazy images, it is highly expected to estimate the depth and airlight in a data-driven instead of handcrafted way.

%To be specific, the data quality heavily depends on the results of the used depth estimation method. Although depth estimation is crucial to hazy image rendering, it is not the only key factor. To generate the favorable hazy images, it is highly expected to estimate the depth and airlight in a data-driven instead of handcrafted way. 

%However, such handcrafted hazy images are probably less informative and inconsistent with the real hazy images, and thus would lead to the domain shift issue, \textit{i.e.}, using the synthetic dataset to train the model which is further applied to the real-world hazy images~\cite{Shao:2020we, YOLY}. 

This work is different from these haze rendering studies in the given aspects. First, our method does not simply treat the haze rendering as a depth or transmission map estimation problem. Instead, we learn the transmission map from the input clean image and the airlight from a given real hazy image. With the learned depth and airlight, our method could generate pleasant results, thus alleviating and even avoiding the domain shift issue. Second, our method estimates the transmission map and airlight in an unsupervised way. It does not require paired images, thus avoiding labor-intensive data collection. Third, our method could generate hazy images in a controllable way. In other words, the haze densities and airlights of the rendered images could be explicitly specified by changing the value of $\alpha$ and the exemplar. 

%To alleviate above problems, we propose a novel hazy image rendering method termed HazeGEN. HazeGEN is remarkably different from existing hazy images generation approaches in the following two aspects. First, most existing physical-based methods do not consider the importance of the airlight. To be specific, almost all of these methods adopt the value of airlight by randomly sample, while HazeGEN adaptively learn the airlight from the real hazy images, which makes the generated images more plausible. Second, HazeGEN has less limits during the usage and can be adopted to every images. Most of the existing generation methods have several requirements to the generated scenes. Device-based methods can not render existing clean images and depth estimated based method could not deal with the scene which is different from the training sets. Our proposed model trained our methods in an unsupervised way, which avoids above problems.   

\begin{figure*}[!t]
	\begin{center}
		\includegraphics[width=1\textwidth]{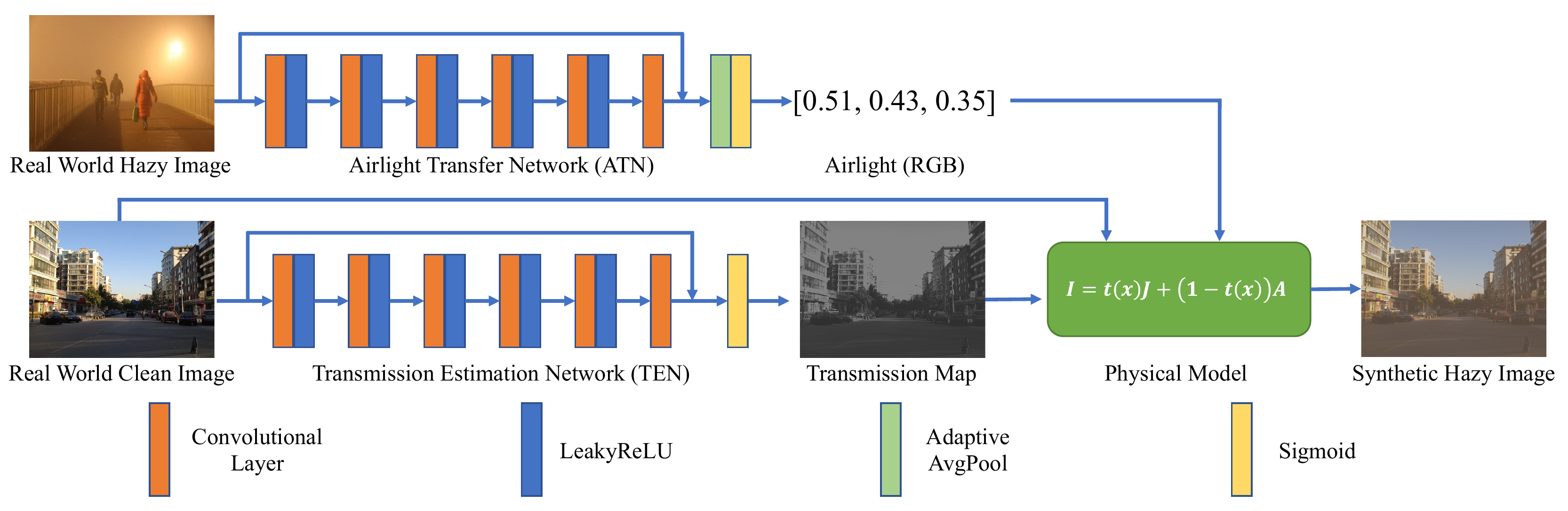}
%		width=\m_width\textwidth, height=\a_height
	\end{center}
	\caption{\label{Figure:Net} The framework of HazeGEN which consists of two modules, \textit{i.e.}, Transmission Estimation Network (TEN) and the Airlight Transfer Network (ATN). In short, TEN and ATN  are designed to implement the  structure similarity prior and airlight consistency prior, respectively.}
\end{figure*}

\section{Method}

In this section, we elaborate on the proposed HazeGEN method which estimates the depth and airlight by leveraging two new priors to build two modules, \textit{i.e.}, structure similarity prior driven Transmission Estimation Network (TEN) and airlight consistency prior driven Airlight Transfer Network (ATN). For clarity, we will first introduce the proposed loss function and then elaborate on implementation details of the modules.

%As illustrated in Fig.~\ref{Figure:Net}. HazeGEN consists of a Transmission Estimation Network (TEN) and an Airlight Transformation Network (ATN). For clarity, we will first introduce the proposed loss function and then elaborate the implementation details of each subnetwork.
\subsection{The objective Function}
As discussed above, almost all existing haze synthesis methods have suffered from the following two issues, 1) to estimate the depth or equivalently the transmission map, they need auxiliary information including but not limited to paired data; and 2) almost all of them ignore the importance of the airlight and determine it by randomly sampling. 
 
To address these two issues, we proposed a novel hazy image rendering method termed HazeGEN. As illustrated in Fig.~\ref{Figure:Net}, HazeGEN consists of two knowledge-driven modules, \textit{i.e.}, TEN and ATN. To be specific, for a given input clean image $x$, HazeGEN first feeds it into TEN to predict the transmission map  with a controllable density parameter $\alpha$. Meanwhile, we pass a real-world hazy image (called exemplar) $y$ through ATN to obtain the corresponding airlight $A$. With $t^{\alpha}(x)$ and $A$, the haze could be rendered into the clean image $x$ to obtain the hazy image $z$ through the physical model. To jointly train the above subnetworks, our loss function is in the following form:
\begin{equation}
	\label{eq:all}
	\mathcal{L}=\lambda_{S}\mathcal{L}_{S}+\lambda_{A}\mathcal{L}_{A} + \lambda_{adv}\mathcal{L}_{adv},
\end{equation}
where $\lambda_{S}$, $\lambda_{A}$ and $\lambda_{adv}$ are nonnegative factors which balance the importance of the three terms. In our experiments, we simply set $\lambda_{S}=1.0$, $\lambda_{A}=0.1$, and $\lambda_{adv}=1.0$. $\mathcal{L}_{S}$, $\mathcal{L}_{A}$, and $\mathcal{L}_{adv}$ are designed to for the following goals. To be exact, $\mathcal{L}_{S}$ is used to achieve the structure similarity prior between the estimated depth and the input clean image $x$, $\mathcal{L}_{A}$ enforces the airlight consistency on the rendered image $z$ and the hazy exemplar $y$, and $\mathcal{L}_{adv}$ is used to improve the data quality. In the following, we will elaborate on these terms one by one.

\textbf{Structure Similarity Prior.}
As illustrated in Fig.~\ref{Fig:BasicIdea}, there exists the structure similarity between the depth map and the input clean image, and thus we proposed a similarity prior $\mathcal{L}_{S}$ for TEN:
\begin{equation}
	\label{eq:T}
	\mathcal{L}_{S}=\lambda_{e}\mathcal{L}_{e}+\lambda_{l}\mathcal{L}_{l} + \lambda_{smooth}\mathcal{L}_{smooth},
\end{equation}
where $\lambda_{e}$, $\lambda_{l}$, and $\lambda_{smooth}$ are three balanced factors which are fixed to $0.25$, $0.25$, and $1.0$, respectively. Motivated by our observation, we refer the structure to two parts. The first part is high frequency (\textit{i.e.}, contour) which is formulated into $\mathcal{L}_{e}$ via 
\begin{equation}
	\label{eq:edge}
	\mathcal{L}_{e}=\|x-t(x)\|_{1}. 
\end{equation}

We refer the second structure to the luminance information (\textit{e.g.}, brightness and contrasts) which could be achieved by:   
\begin{equation}
	\label{eq:lum}
	\mathcal{L}_{l}=\frac{(2 x_\mu t_{\mu} + c_1) (2\sigma_{xt}+c_2)}{(x_\mu^2 + t_{\mu}^2 + c_1) (\sigma_x^2 + \sigma_t^2 +c_2)},
\end{equation}
where $x_\mu$ and $t_{\mu}$ are the mean of $x$ and $t(x)$, respectively. $\sigma_x$ and $\sigma_{t}$ are the variance of $x$ and $t(x)$, respectively. $\sigma_{xt}$ is the covariance of $x$ and $t(x)$. $c_1=(0.01L)$ and $c_2=(0.03L)$ are the constant used to stabilize the model, where $L$ the possible maximal pixel value. Note that, $\mathcal{L}_{l}$ is actually the well-known SSIM metric which will also preserve the contour information.  

The loss $\mathcal{L}_{smooth}$ is designed to encourage the transmission map to be locally smooth by enforcing an $l_1$ penalty on the transmission gradients. Formally, 
\begin{equation}
	\label{eq:smooth}
	\mathcal{L}_{smooth}=\frac{1}{N}\sum_{ij}|\nabla_r t_{ij}|e^{- \|\nabla_r t_{ij} \|} + |\nabla_c t_{ij}|e^{- \|\nabla_c t_{ij} \|},
\end{equation}
where $t_{ij}$ denotes the value at the $i$-th row and the $j$-th column of $t(x)$, and $\nabla_r$/$\nabla_c$ denotes the gradient along the row/column. 

\textbf{Airlight Consistency Prior.}
In order to generate a favorable hazy image $z$, we transfer the airlight from the given exemplar  $y$ into the clean image $x$ through ATN with the following loss:
\begin{equation}
	\label{eq:p}
	\mathcal{L}_{A}=\lambda_{s}\mathcal{L}_{s}+\lambda_{c}\mathcal{L}_{c},
\end{equation}
where both the parameters $\lambda_{s}$ and $\lambda_{c}$ are fixed to $0.1$ through the experiments. The loss $\mathcal{L}_{s}$ is the so-called  perceptual loss~\cite{ploss}, which enforces $z$ and $y$ are similar at high-level features. As our goal is to transfer the airlight from $y$ into $x$ while keeping the content of $x$ unchanged, the visual quality would be degraded if only $\mathcal{L}_{s}$ is used. Hence, we design $\mathcal{L}_{c}$ to keep the content consistency by enforcing $z$ similar to $x$ at the low-level features. Formally, 
\begin{equation}
	\label{eq:ls}
	\mathcal{L}_{s}=\frac{1}{N_{H}}\sum^{N_{H}}_{i}\| f_{i}(y) - f_{j}(z) \|^{2}_{2},
\end{equation}
and 
\begin{equation}
	\label{eq:lc}
	\mathcal{L}_{c}=\frac{1}{N_{L}}\sum^{N_{L}}_{i}\| f_{i}(x) - f_{j}(z) \|^{2}_{2},
\end{equation}
where $f_{i}(\cdot)$ is the $i$-th layer of VGG16~\cite{vgg}. For $\mathcal{L}_{s}$, we compute loss at layer relu1\_2, relu2\_2, relu3\_3, and relu4\_3 of VGG16. For $\mathcal{L}_{c}$, we compute loss at layers relu3\_3.

%In order to transfer airlight of $y$ into the synthetic hazy image $z$, we use perceptual loss to update the ATN. It could be work owing to the structure design of ATN. Specifically, the output of ATN is a three-channel's RGB value, it could only influence the color of synthetic hazy images $z$ according to the atmospheric scattering model. To minimize the given perceptual loss, the ATN will try its best to get the airlight information, which lets the ATN transfers the airlight from $y$ to $z$. As indicated in~\cite{ploss}, higher layers in vgg16~\cite{vgg} tend to capture perceptual information and lower layers tend to capture texture information. To get the airlight information, we adopt a vgg16 model pre-trained on image classification task. Then, we minimize the higher layer features calculated from the $y$ and $z$. To keep the texture information in the original clean image $x$, we also minimize the lower layer features calculated from the $x$ and $z$. Mathematically, the perceptual loss function is presented as follow

To further enhance the reality of the synthesized hazy images, we also employ an adversarial loss on the exemplar $y$ and obtained hazy image $z$. Specifically, we treat $y$ as the sample from the ``real'' domain, and $z$ as the sample from the ``fake'' domain. In mathematically, 
\begin{equation}
	\label{eq:adv}
	\mathcal{L}_{adv}=-\frac{1}{N}\left(\sum_{y}\log\left(D\left(y\right)\right) + \sum_{z}\log\left(1 - D\left(z\right)\right)\right),
\end{equation}
where $D(\cdot)$ is a discriminator network and we adapt the network architecture used in~\cite{Anonymous:ETAb_elU}.

\subsection{Network Architecture and Implementation}
In this section, we briefly introduce the network architecture and the implementation of our method, and more details have been presented in the supplementary material. As introduced above, our HazeGEN consists of TEN and ATN, where TEN takes a non-downsampling architecture by following~\cite{YOLY} to better keep the details of transmission map. In other words, TEN only consists of convolutional layers and LeakyReLU activation with skip connections. In the final layer, we adopt sigmoid to normalize the output into $[0, 1]$. ATN adopts a similar network structure to TEN. The only difference between them is that ATN adds an adaptive average pooling layer before the sigmoid layer so that the dimension of the output is desired, \textit{i.e.}, from $\mathbb{R}^{3\times m\times n}$ to $\mathbb{R}^{3\times1}$, where $m$ and $n$ are the height and width of the input image, respectively.

\section{Experiments}

In this section, we conduct the following experiments to validate the effectiveness of our method. First, we will carry out qualitative and quantitative studies to show the reality of the hazy image rendered by HazeGEN. Second, we verify the effectiveness of our method in boosting image dehazing approaches. Third, we conduct qualitative experiments to investigate the influences of the haze density parameter $\alpha$  and the haze transferability by ATN. Finally, we carry out an ablation study to verify the role of our objective function. Due to the space limitation, we leave more experimental details and results in the supplementary materials.

\subsection{Experimental Settings}

In this section, we introduce the details of the datasets, baselines, evaluation metrics, and implementation details.
%To verify the effectiveness of our method, we compared the hazy images rendered by HazeGEN with some existing data generation methods. We first briefly introduce the datasets we used for rendering the hazy images and evaluating the rendered images.

\textbf{Datasets:} 
\label{datasets}
To conduct experiments, we use the following five datasets, \textit{i.e.}, one indoor synthetic dataset $\mathcal{D}^{input}_{in}$ and one outdoor synthetic dataset $\mathcal{D}^{input}_{out}$ as the inputs; one real-world indoor hazy dataset $\mathcal{D}^{gt}_{in}$ and one real-world outdoor hazy dataset $\mathcal{D}^{gt}_{out}$ as the ground-truth hazy images; and another real-world hazy dataset $\mathcal{D}^{e}$ as exemplars. To be specific, 
 $\mathcal{D}^{input}_{in}$ consists of the indoor Training Set (ITS) from RESIDE~\cite{Reside}, which contains 1,399 clean indoor images and 13,990 synthetic hazy images gathered from NYU2 dataset~\cite{NYU2} and MiddleBury stereo dataset~\cite{middle}, respectively.
% , whose depth maps are measure by Kinect or structured light.
% For images from NYU2 datasets, the depth map are measured by Kinect. For images from MiddleBury stereo datasets, the depth maps are estimated by the structured light. For each clean image in ITS, RESIDE synthesizes the hazy images with two different hand-crafted airlights and five different haze densities. 
 $\mathcal{D}^{input}_{out}$ consists of the Outdoor Training Set (OTS) from RESIDE, which contains 2,061 clean outdoor images and 72,135 synthetic hazy images collected from the Internet and synthesized by~\cite{Liu:2016ds}, respectively. 
$\mathcal{D}^{e}$ consists of the Real-world Task-driven Testing Set (RTTS) from RESIDE, which contains 4,322 outdoor hazy images collected from the real world. 
$\mathcal{D}^{gt}_{in}$ and $\mathcal{D}^{gt}_{out}$ are I-HAZE~\cite{i-haze} and O-HAZE~\cite{o-haze}, respectively. In short, I-HAZE contains 35 indoor scenes and O-HAZE contains 45 outdoor scenes. In I-HAZE and O-HAZE, the hazy-clean image pairs are generated by the haze machines, which are credible and plausible.

\textbf{Baselines:} 
Although there are several graphics-based methods have been proposed, they have suffered from many issues and heavily rely on the auxiliary information. As it is daunting to obtain the auxiliary information for the datasets used in this work, we compare our neural rendering method with the popular haze generation approach used in RESIDE~\cite{Reside}. In brief, the approach generates the ITS dataset based on the depth map estimated using the Kinect, and synthesizes the OTS dataset relies on pre-trained depth estimated models. Both of them work in a two-stage way, which first estimates the depth map of given hazy images and then synthesizes the hazy images based on the physical model with a random sampling airlight. 

\textbf{Evaluation metrics:} 
It is a challenging and open issue how to quantitatively evaluate the generation results of image hazing due to the following reason. To be exact, haze image rendering aims to simulate the degradation process, which tries to make the quality of images ``worse''. However, most of the existed metrics evaluate how ``well'' an image is, thus leading to the difficulty in using existing image quality evaluation metrics. To address this issue, we adopt Fréchet inception distance (FID) for quantitive comparisons  following~\cite{Li:2021vv}. FID is a metric to evaluate the distance between the feature vectors calculated from real and the generated images. In other words, it could measure the difference between two distributions even though the image contents are different. The smaller the distance, the better the performance. Moreover, 
to validate the effectiveness of our method in image dehazing, we adopt Peak Signal-to-Noise Ratio (PSNR) as the metric.
%https://machinelearningmastery.com/how-to-implement-the-frechet-inception-distance-fid-from-scratch/

\textbf{Training details:} 
\label{settings}
We implement experiments on one NVIDIA GeForce RTX 2080Ti GPU in PyTorch and employ the ADAM optimizer~\cite{Adam} with the learning rate of 0.001 to train HazeGEN. For training, we use $\mathcal{D}^{input}_{in}$ and $\mathcal{D}^{input}_{out}$ as the training set and $\mathcal{D}^{e}$ as exemplars. The maximal iteration number is set to 100. In each iteration, we train our model with 100 batches with the batch size of 32 and patch size of 128. Note that, our method works in an unsupervised way, and the compared methods following their recommended experimental protocols.

\begin{table*}
%\begin{small}
\centering
\caption{Reality analysis on the synthesized hazy image. The bold number indicates the best result.}
\label{Table:reality}
%\begin{footnotesize}
\begin{tabular}{c ccc  ccc}
\toprule
\multicolumn{1}{c}{\multirow{2}{*}{Metrics}} & \multicolumn{3}{c}{Indoor Scene}  & \multicolumn{3}{c}{Outdoor Scene} \\
\cline{2-7}
%\midrule
 & Clean & ITS & Ours & Clean & OTS & Ours \\ 
\midrule
FID & 260.94 & 254.55 & \multicolumn{1}{c}{\textbf{203.34}} & 250.67 & 236.06 & \textbf{234.48}\\
Subjective Evaluation & - & 15.66\% & \multicolumn{1}{c}{\textbf{84.34\%}} & - & 25.66\% & \textbf{74.34\%}\\
Image Dehazing & - & 14.34 & \multicolumn{1}{c}{\textbf{14.69}} & - & 15.02 & \textbf{15.62}\\
\bottomrule
\end{tabular}
\end{table*}

\subsection{Reality Analysis}
Reality is one of the most important metrics to evaluate the quality of the generated hazy images. To measure the degree of reality, we use $\mathcal{D}^{e}$ as exemplar, then synthesize haze into the clean images of $\{\mathcal{D}^{input}_{in}, \mathcal{D}^{input}_{out}\}$ into hazy ones. With the obtained  $\{\mathcal{D}^{r}_{in},\mathcal{D}^{r}_{out}\}$, we carry experiments to measure the reality of rendered images from the following aspects, \textit{i.e.}, quantitive comparisons, qualitative comparisons, and subjective evaluations. It is worth noting that the major differences between ITS (OTS) and ours lie on the haze densities and airlight, and the contents of them are the same.

\textbf{Quantitive Comparisons:} To quantitively evaluate the reality of the generated images, we adopt FID as the metric. As shown in  Table.~\ref{Table:reality}, the images rendered by HazeGEN are much closer to the real-world hazy images for both the indoor and outdoor scenes. 

\textbf{Qualitative Comparisons:} We also conduct qualitative comparisons in Fig.~\ref{Figure:1}. As shown, HazeGEN remarkably outperforms the popular haze synthesis method used in RESIDE. To be specific, the baseline  produces some artifacts on the edge and persons caused by the errors in depth estimation. In contrast, benefiting from the proposed structure similarity prior, our method avoids such an issue and obtains more favorable results.

\textbf{Subjective Evaluations:} 
%Though FID could measure the distance between two distributions, however, only this metrics may not enough to handle every situation. 
To further evaluate the visual quality of the rendered images, we conduct a user study to measure the subjective quality by following~\cite{Zhang:2020up}. To be specific, we randomly select 10 images from indoor and outdoor scenes and thus yield 20 images in total for comparisons. We let users select the best results in terms of image reality. As shown in the second row of Table~\ref{Table:reality}, HazeGEN performs better in 84.34\% indoor cases and 74.34\% outdoor cases, which further verifies the reality of the proposed method.

\subsection{Image Dehazing} 
As discussed in Introduction, haze rendering could be helpful in improving the performance of the image dehazing approaches. To verify the effectiveness of our method in boosting image dehazing, we use $\mathcal{D}^{gt}_{in}$ and $\mathcal{D}^{gt}_{out}$ as exemplar for $\mathcal{D}^{input}_{in}$ and $\mathcal{D}^{input}_{out}$, respectively. With the obtained hazy image, we use it to train the dehazing network~\cite{AOD-Net} which further is tested using the real-world datasets ($\mathcal{D}^{gt}_{in}$ and $\mathcal{D}^{gt}_{out}$). As a baseline, we also train the model~\cite{AOD-Net} on ITS (OTS), then test it on $\mathcal{D}^{gt}_{in}$ ($\mathcal{D}^{gt}_{out}$). As shown in Table~\ref{Table:reality}, one could find that the dehazing models trained on our rendered images perform better than the baselines. It indicates that HazeGEN could boost the image dehazing on real-world datasets, thus alleviating the domain-shift issue thanks to the utilization of the airlight consistency prior.

\begin{figure*}[!t]
	\begin{center}
\subfigure{  
\includegraphics[width=2.15cm,height=1.6125cm]{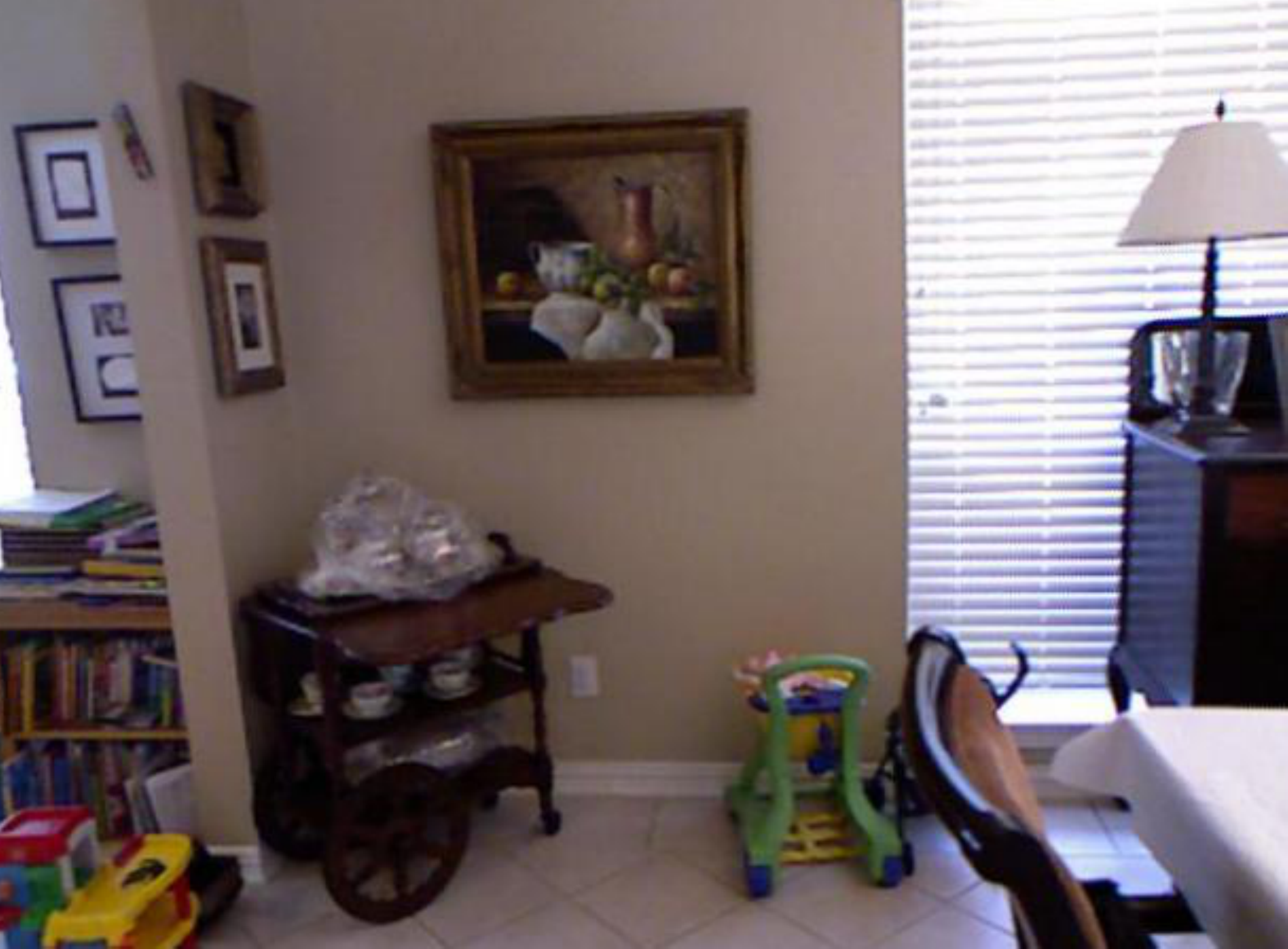}} 
\subfigure{  
\includegraphics[width=2.15cm,height=1.6125cm]{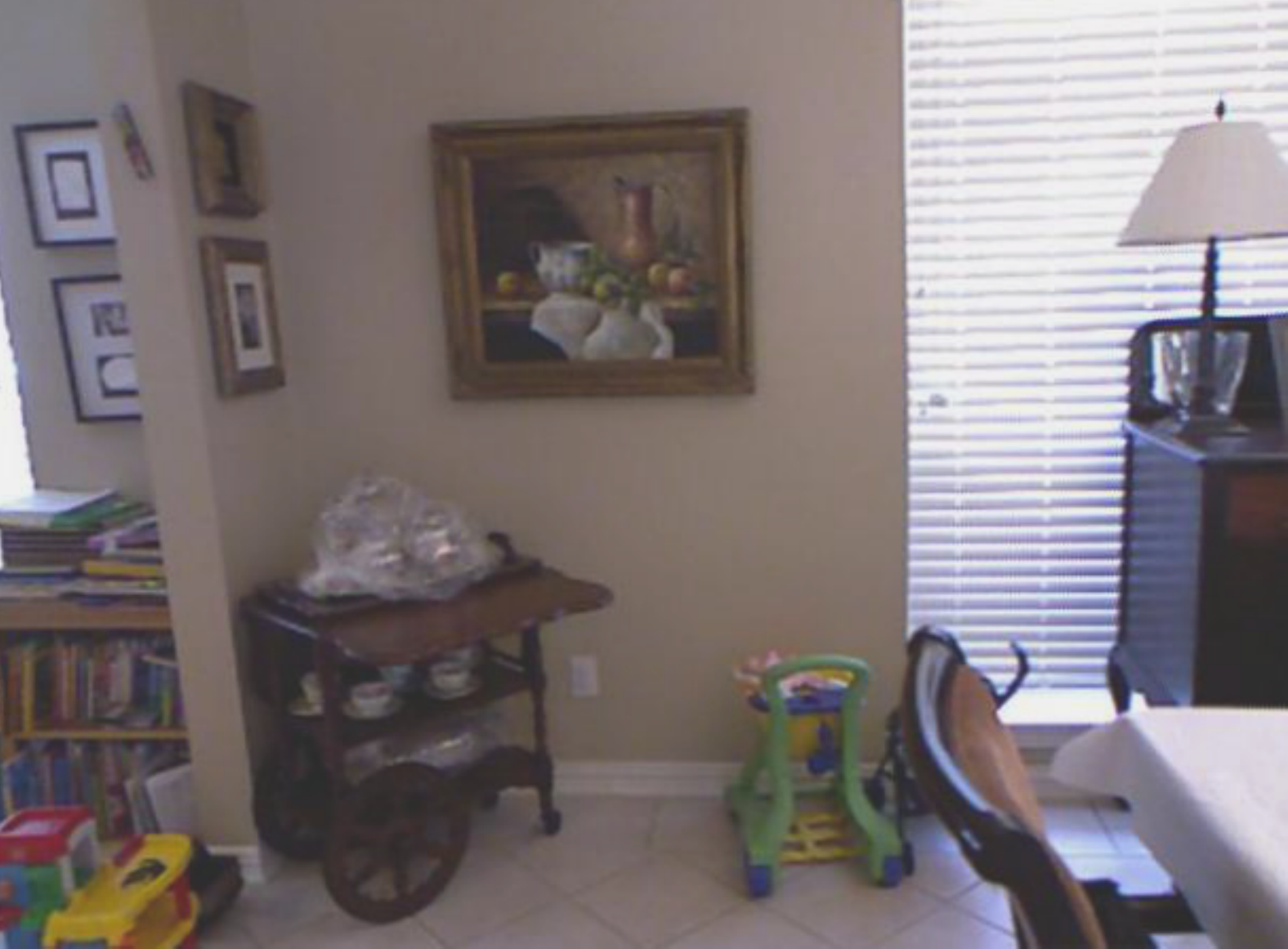}} 
\subfigure{  
\includegraphics[width=2.15cm,height=1.6125cm]{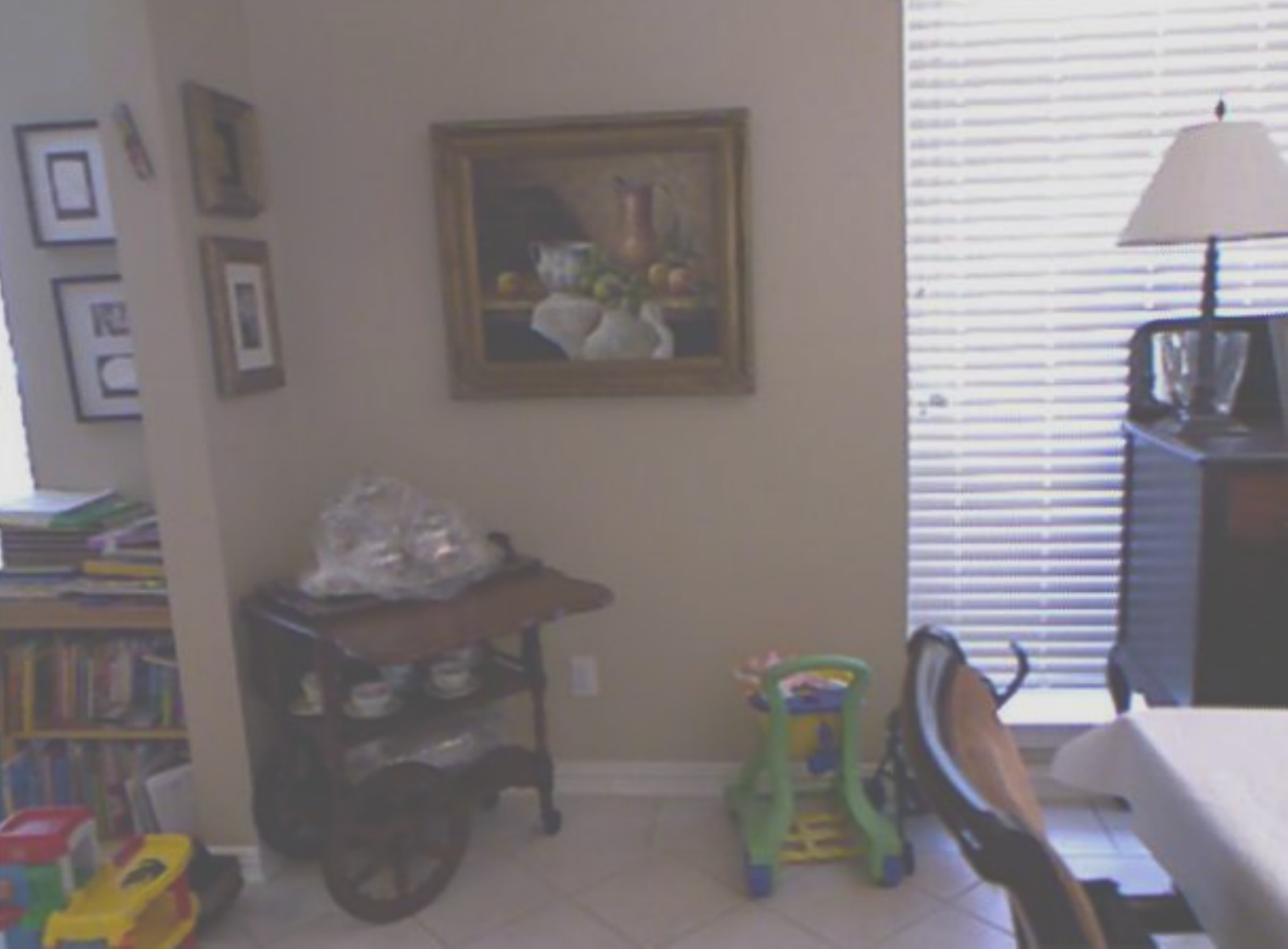}} 
\subfigure{  
\includegraphics[width=2.15cm,height=1.6125cm]{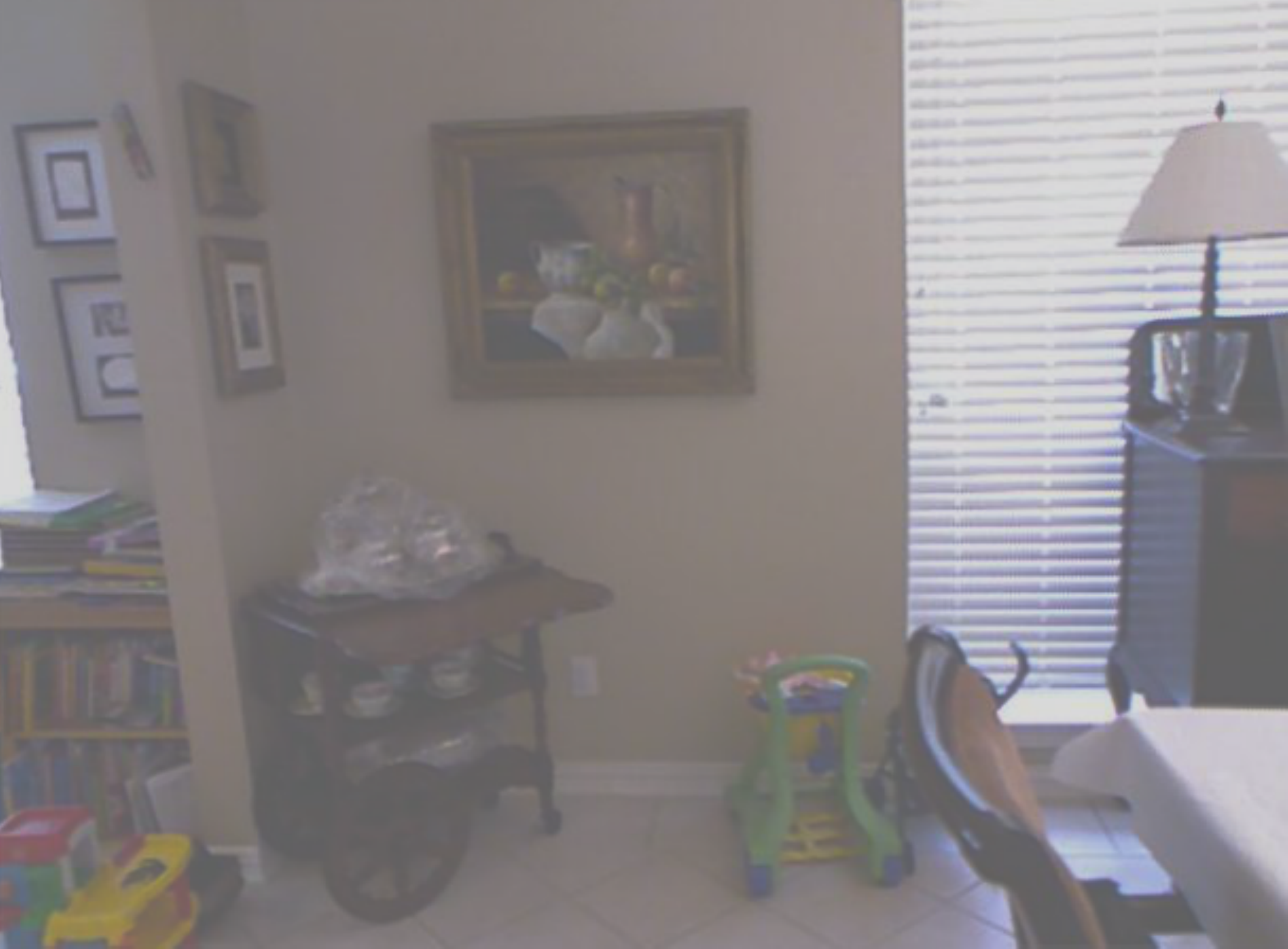}} 
\subfigure{  
\includegraphics[width=2.15cm,height=1.6125cm]{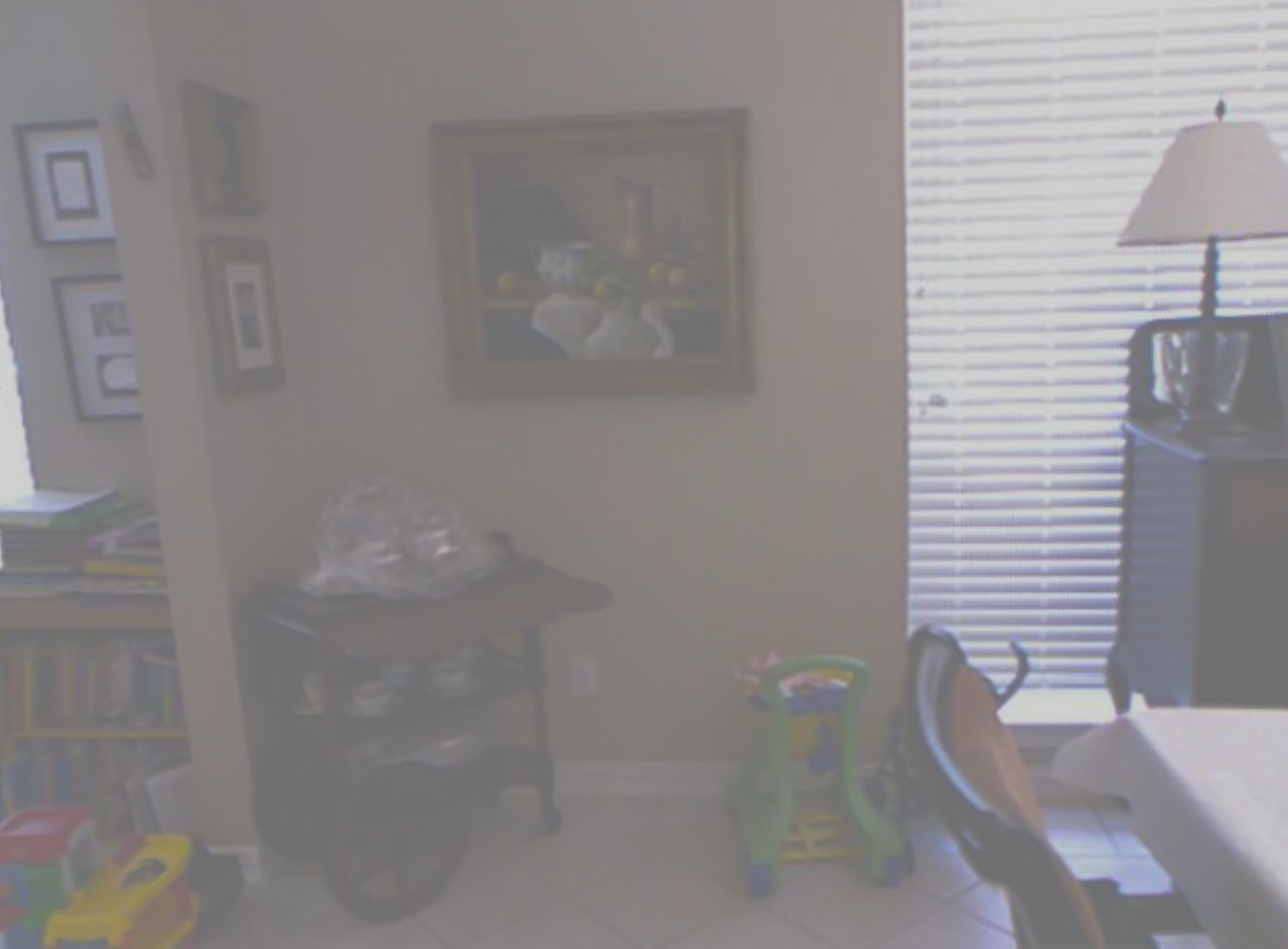}} 
\subfigure{  
\includegraphics[width=2.15cm,height=1.6125cm]{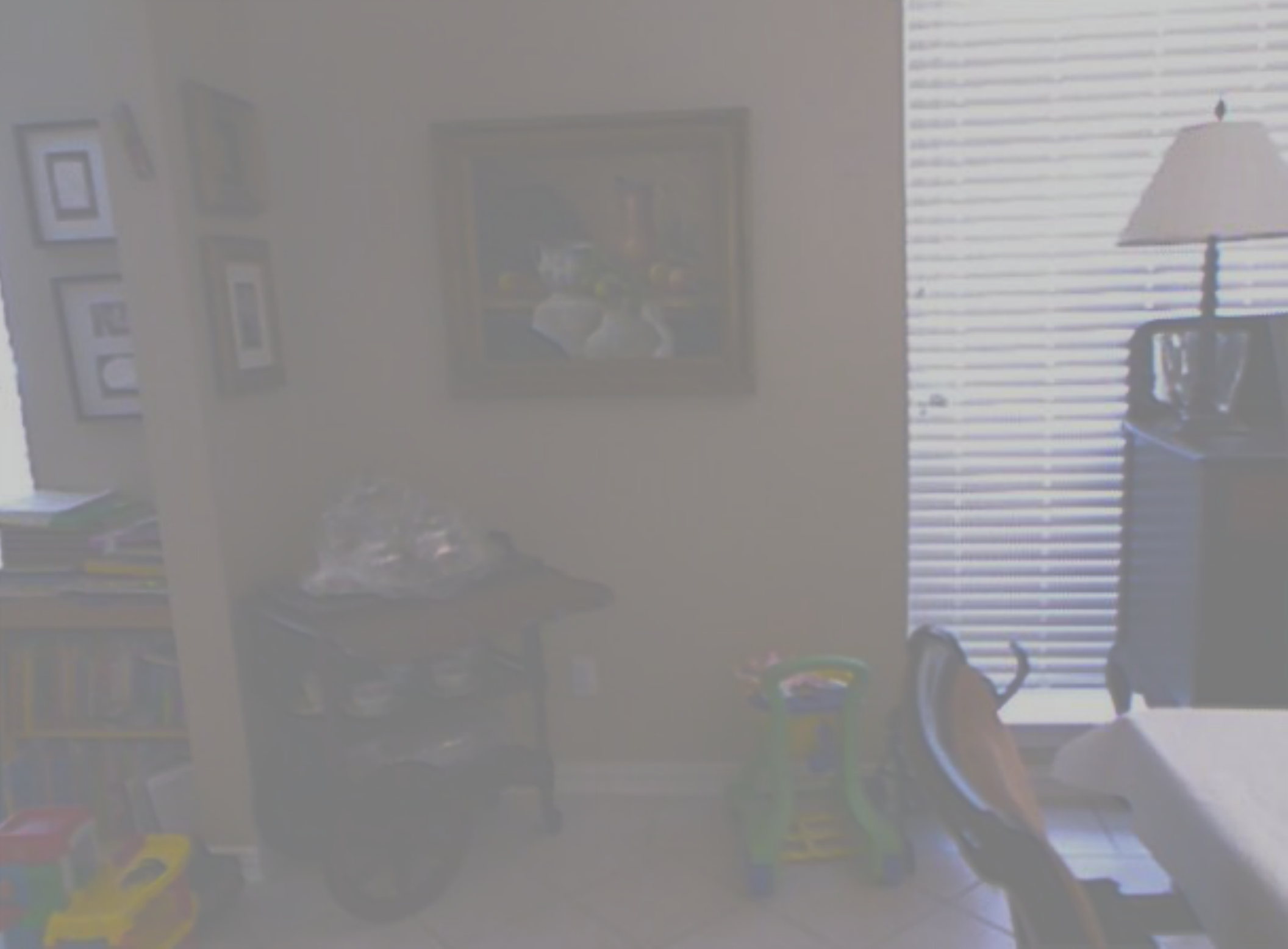}} 

\vspace{-0.2cm}
\setcounter{subfigure}{0}

\subfigure[clean image]{  
\includegraphics[width=2.15cm,height=1.6125cm]{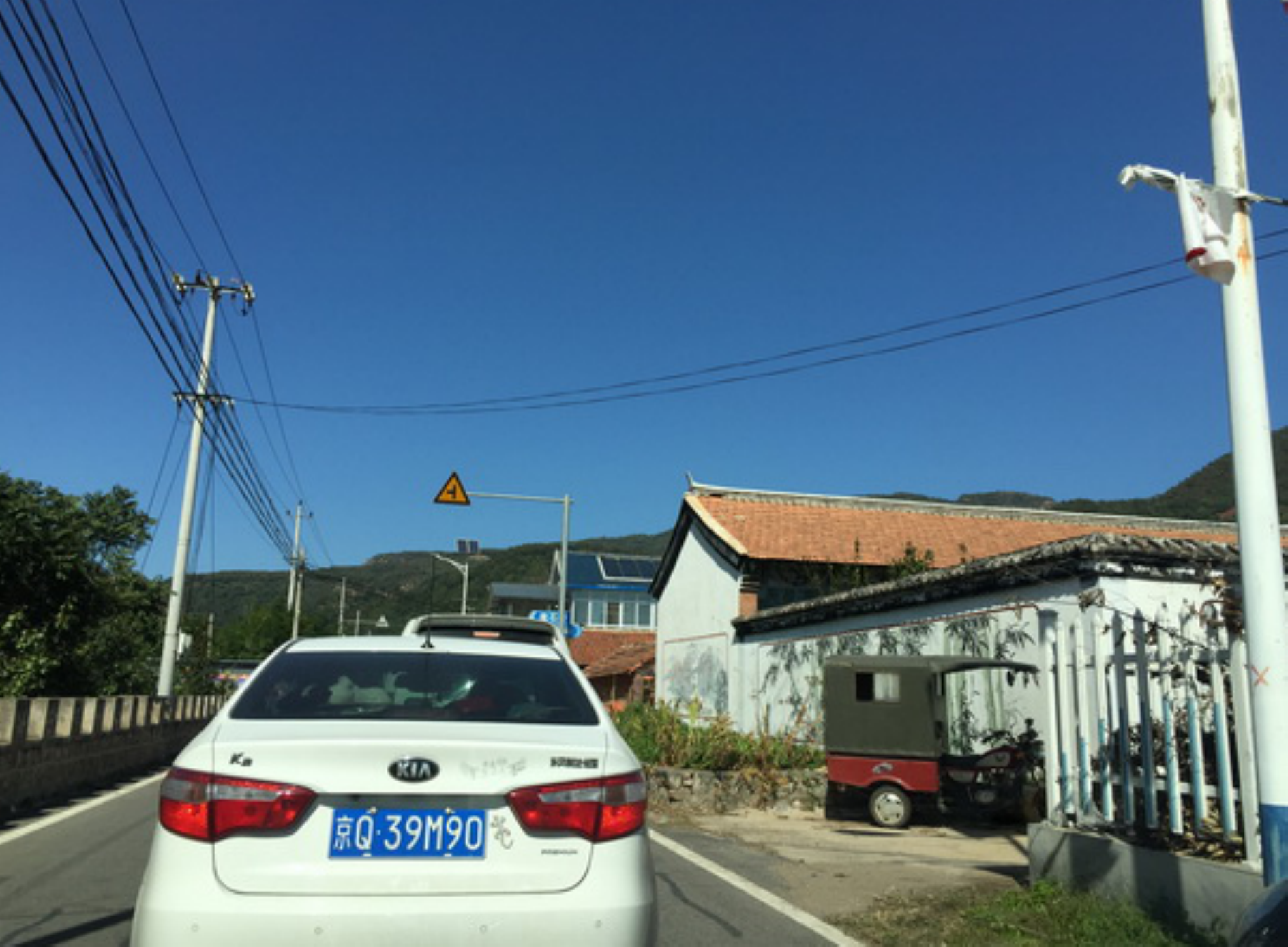}} 
\subfigure[$\alpha = 0.2$]{  
\includegraphics[width=2.15cm,height=1.6125cm]{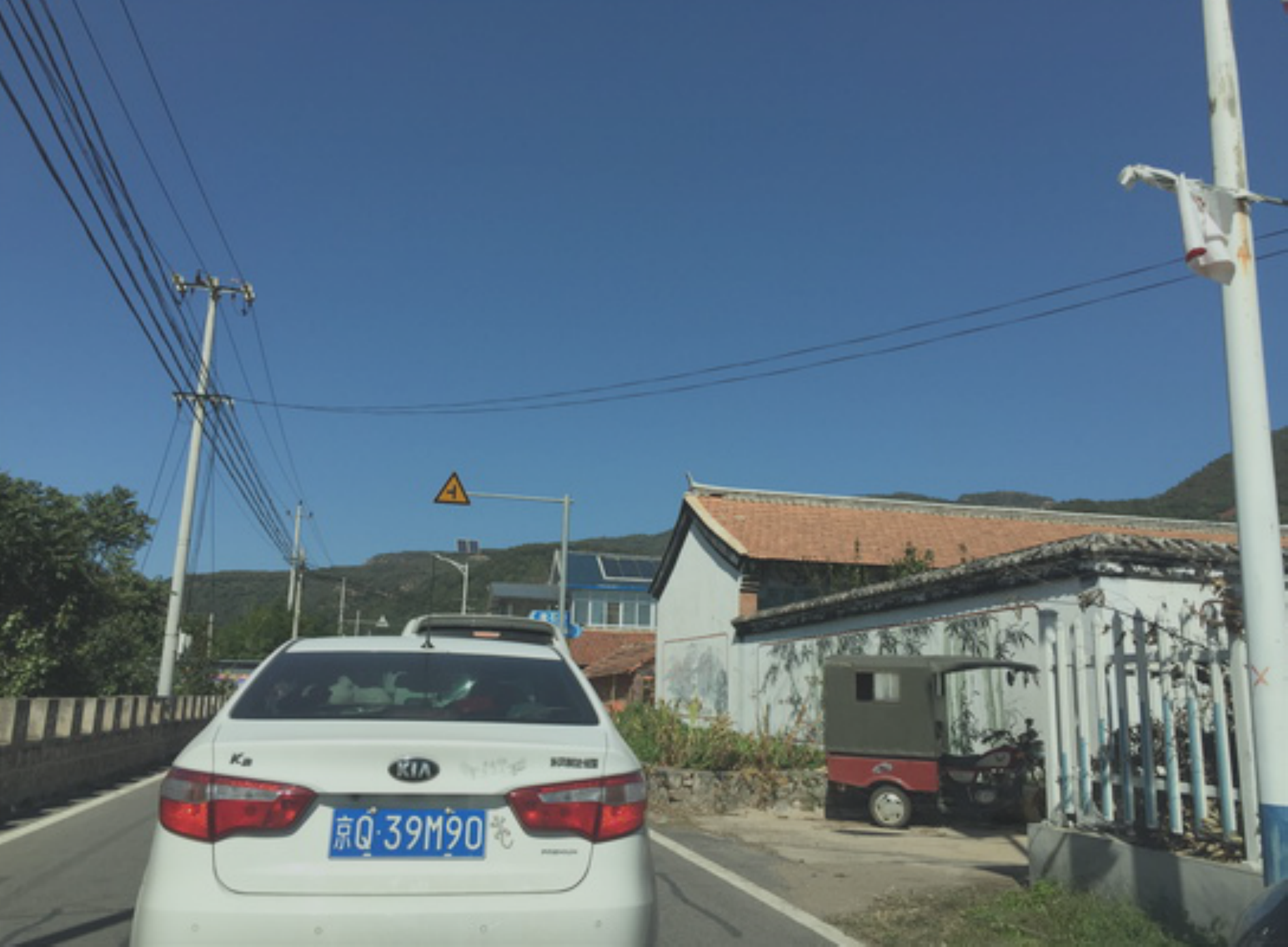}} 
\subfigure[$\alpha = 0.4$]{  
\includegraphics[width=2.15cm,height=1.6125cm]{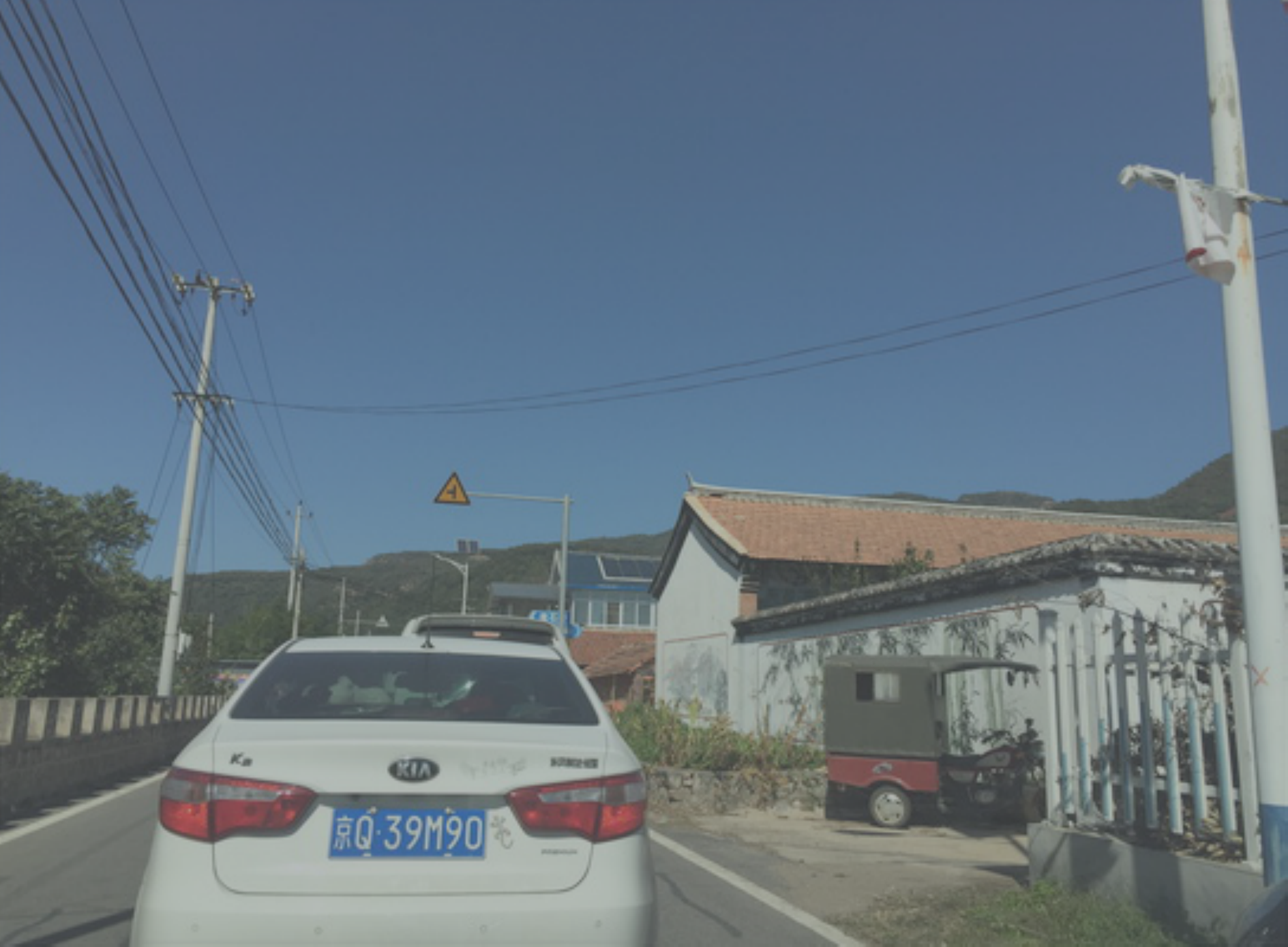}} 
\subfigure[$\alpha = 0.6$]{  
\includegraphics[width=2.15cm,height=1.6125cm]{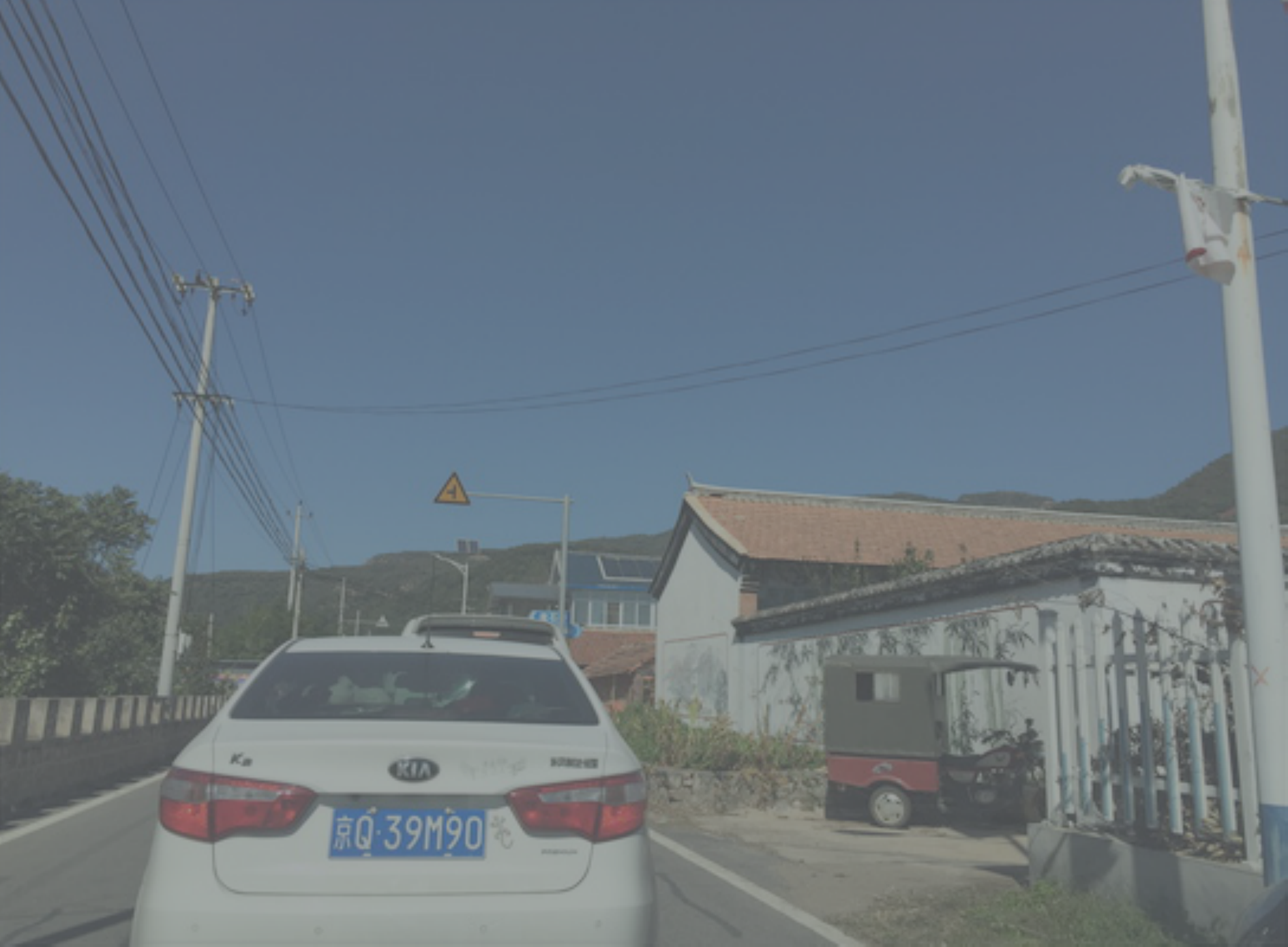}} 
\subfigure[$\alpha = 0.8$]{  
\includegraphics[width=2.15cm,height=1.6125cm]{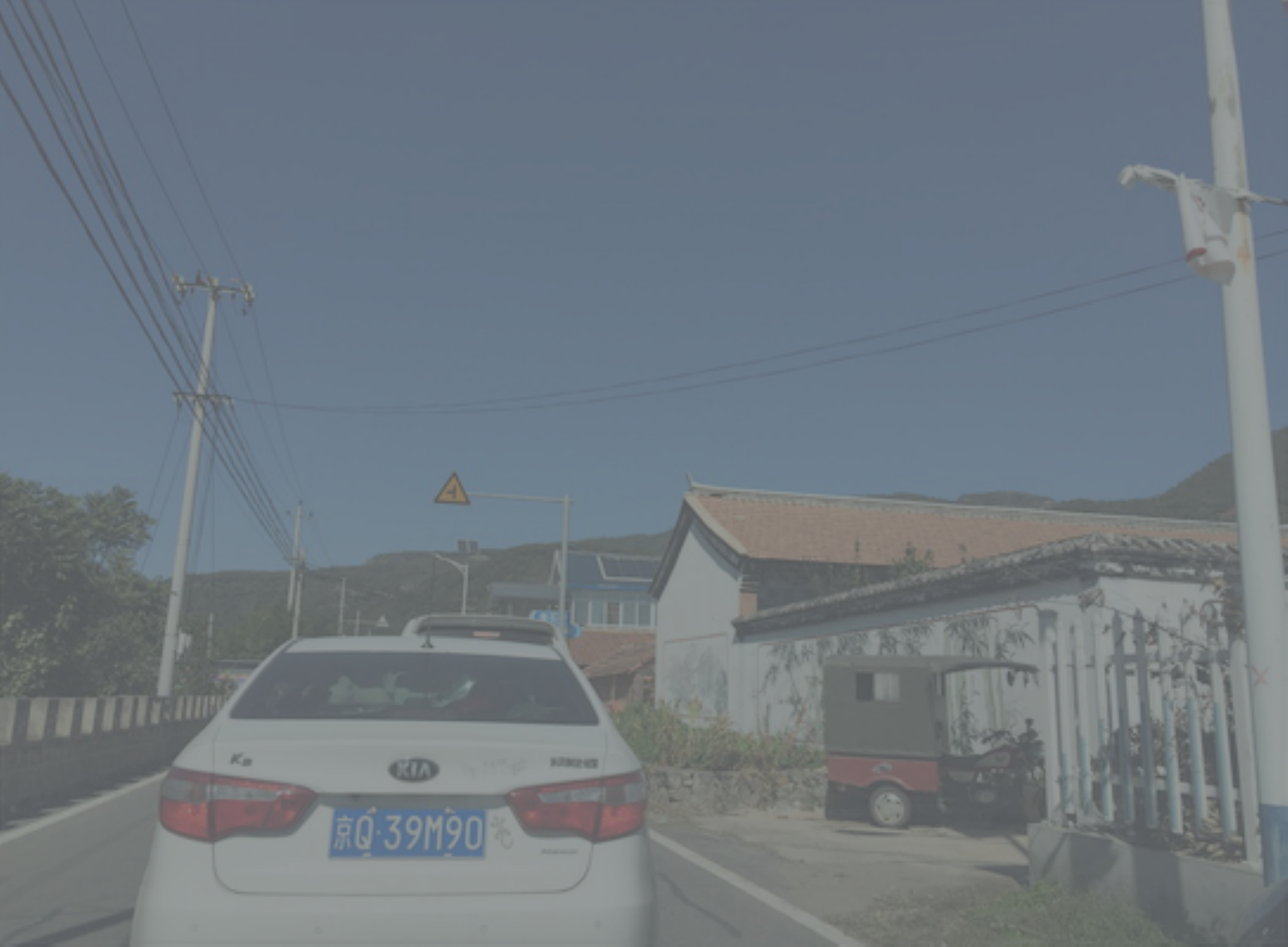}} 
\subfigure[$\alpha = 1.0$]{  
\includegraphics[width=2.15cm,height=1.6125cm]{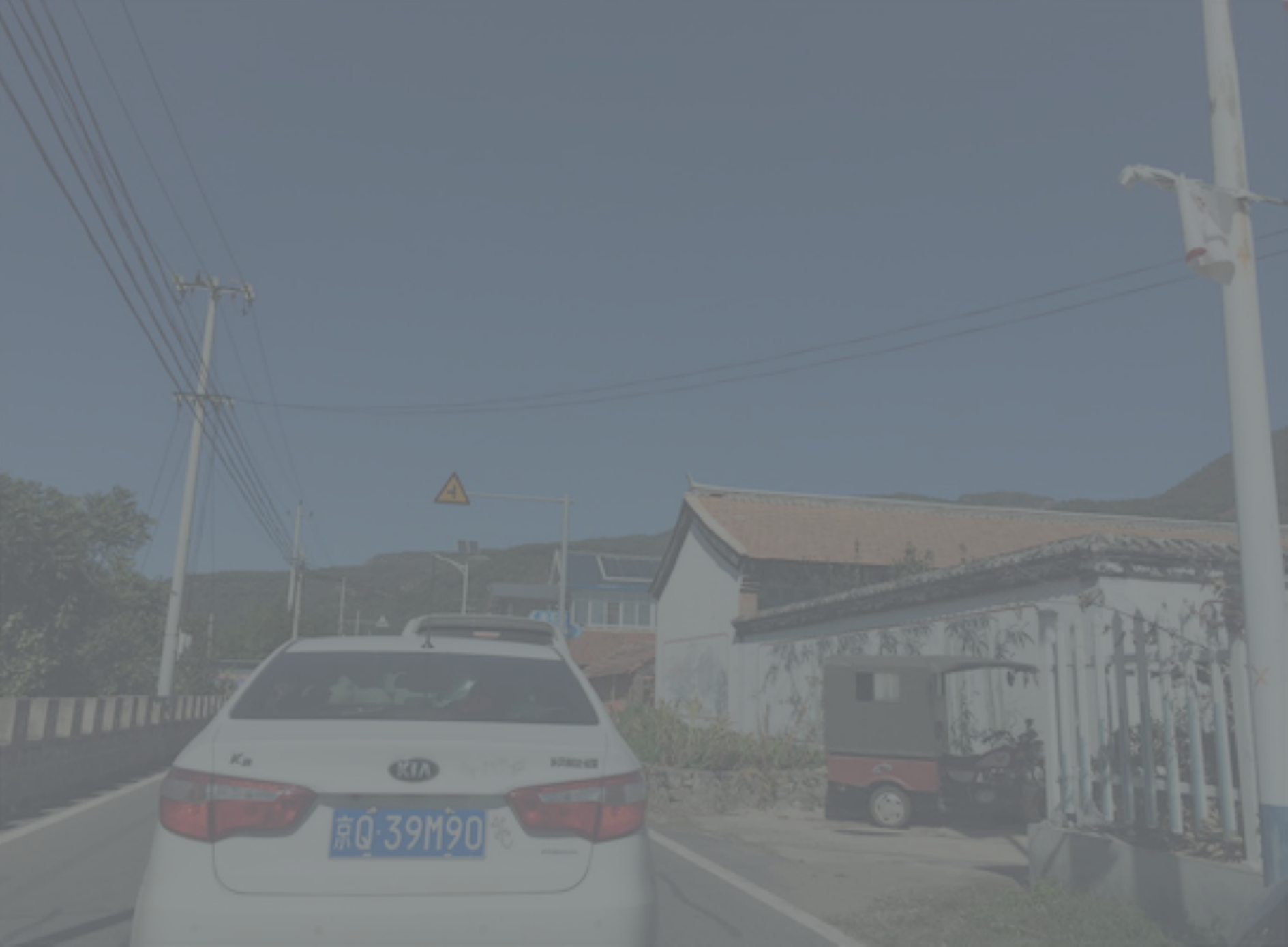}} 
	\end{center}
	\caption{\label{Figure:Bata} Qualitative results of HazeGEN with different scattering coefficients $\alpha$ in the same scene. From the figure, one could find that the densities of the haze increase with the increase of $\alpha$. This verifies that our method could work in a controllable manner. }
\end{figure*}

\subsection{Controllable Multi-density Haze Rendering}

As discussed in the above sections, HazeGEN is a controllable model, \textit{i.e.}, HazeGEN could control the density of haze by adjusting the hyper-parameter $\alpha$. To explore the influences of the hyper-parameter $\alpha$, we render hazy images with different $\alpha$ from 0.0 to 1.0 with a gap of 0.2 in both indoor and outdoor scenes. From  Fig.~\ref{Figure:Bata}, one could observe that the density of the haze in the rendered images is gradually heavier with increasing $\alpha$. This shows that a larger $\alpha$ leads to a larger transmission map, \textit{i.e.}, there are fewer rays from the objects and more airlight come into the scene, thus making heavier haze. 

\begin{figure*}[!t]
	\begin{center}
\subfigure{  
\includegraphics[width=1.8cm,height=1.5cm]{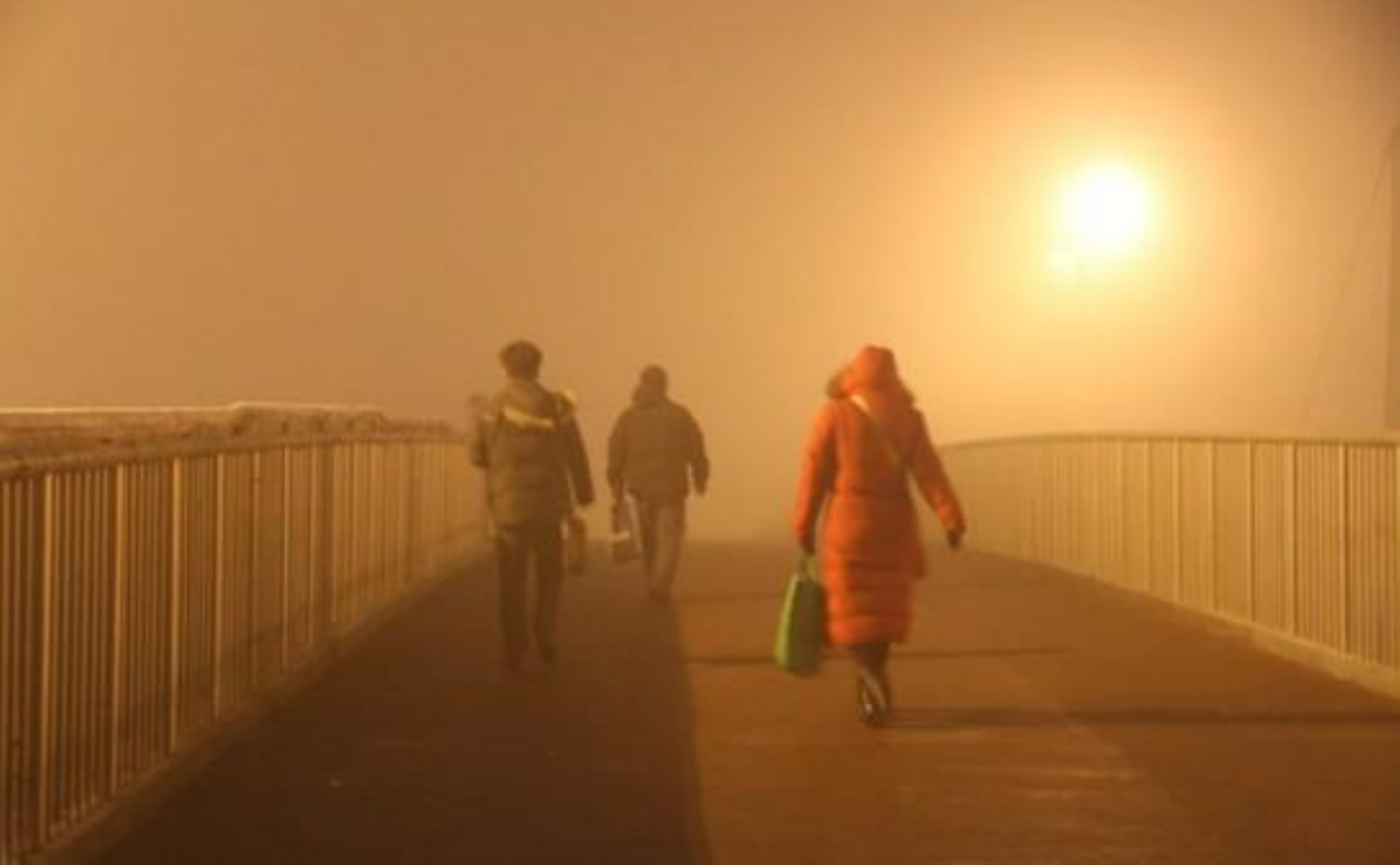}} 
\subfigure{  
\includegraphics[width=1.8cm,height=1.5cm]{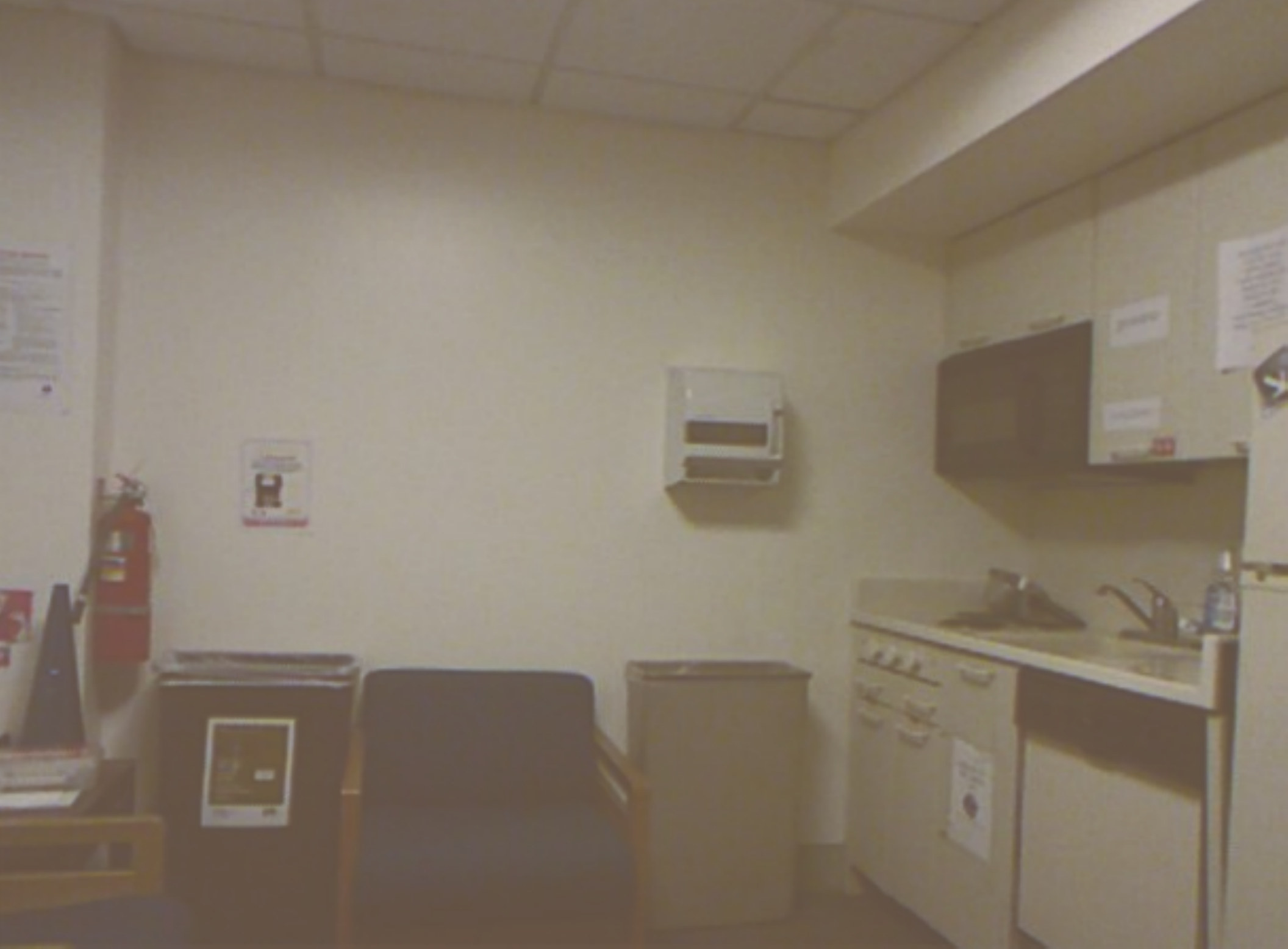}} 
\subfigure{  
\includegraphics[width=1.8cm,height=1.5cm]{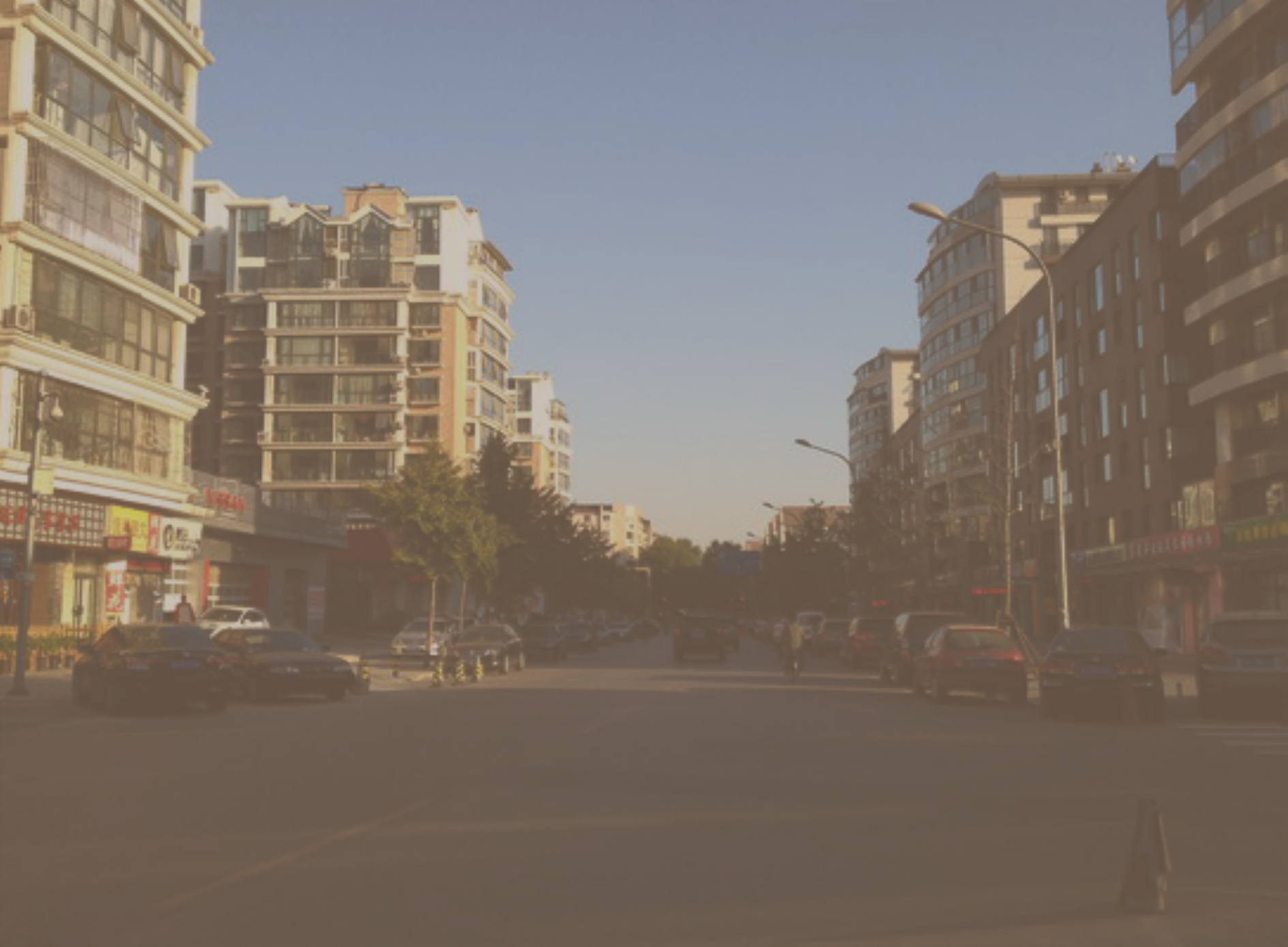}} 
\subfigure{  
\includegraphics[width=1.8cm,height=1.5cm]{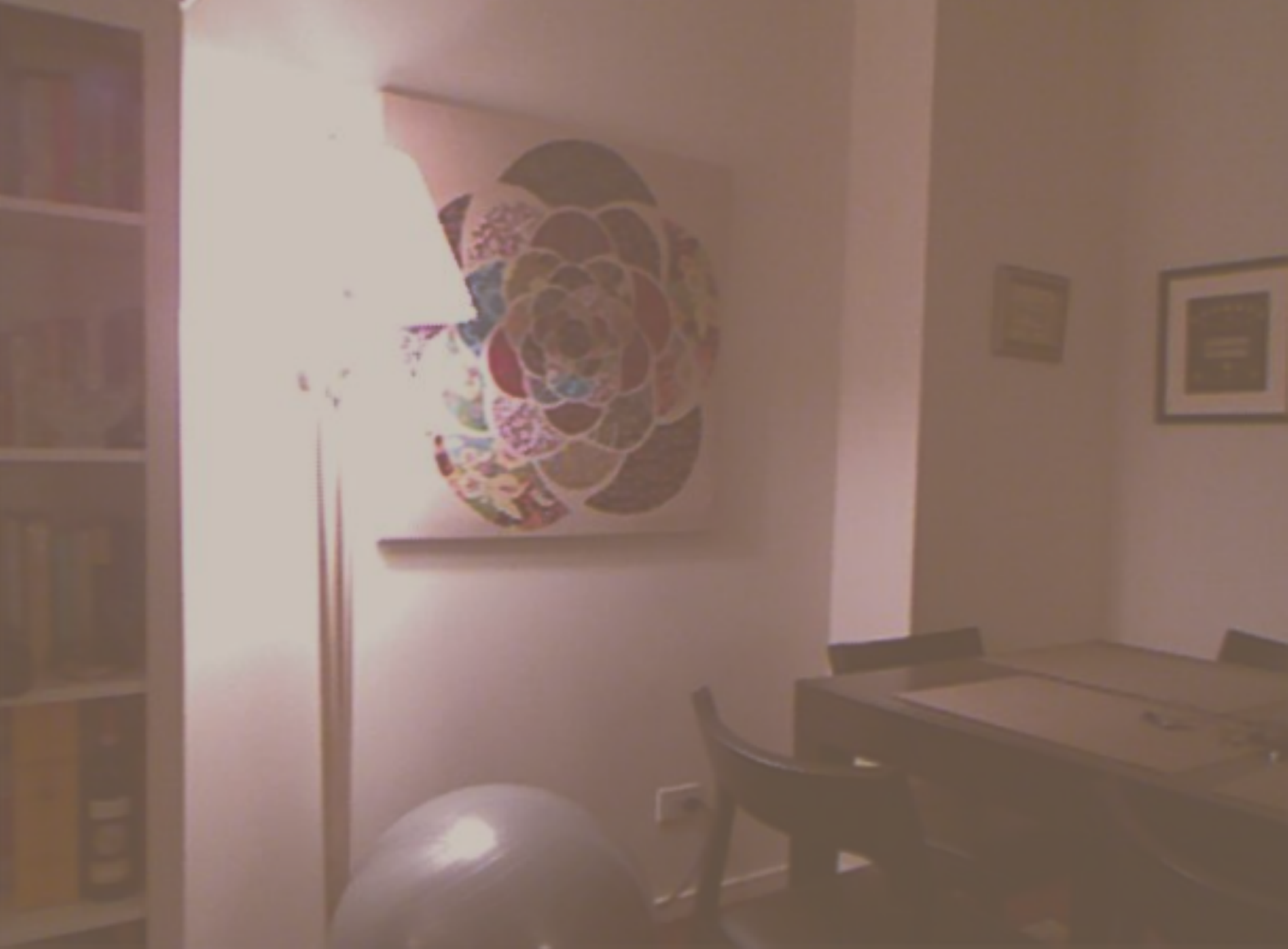}} 
\subfigure{  
\includegraphics[width=1.8cm,height=1.5cm]{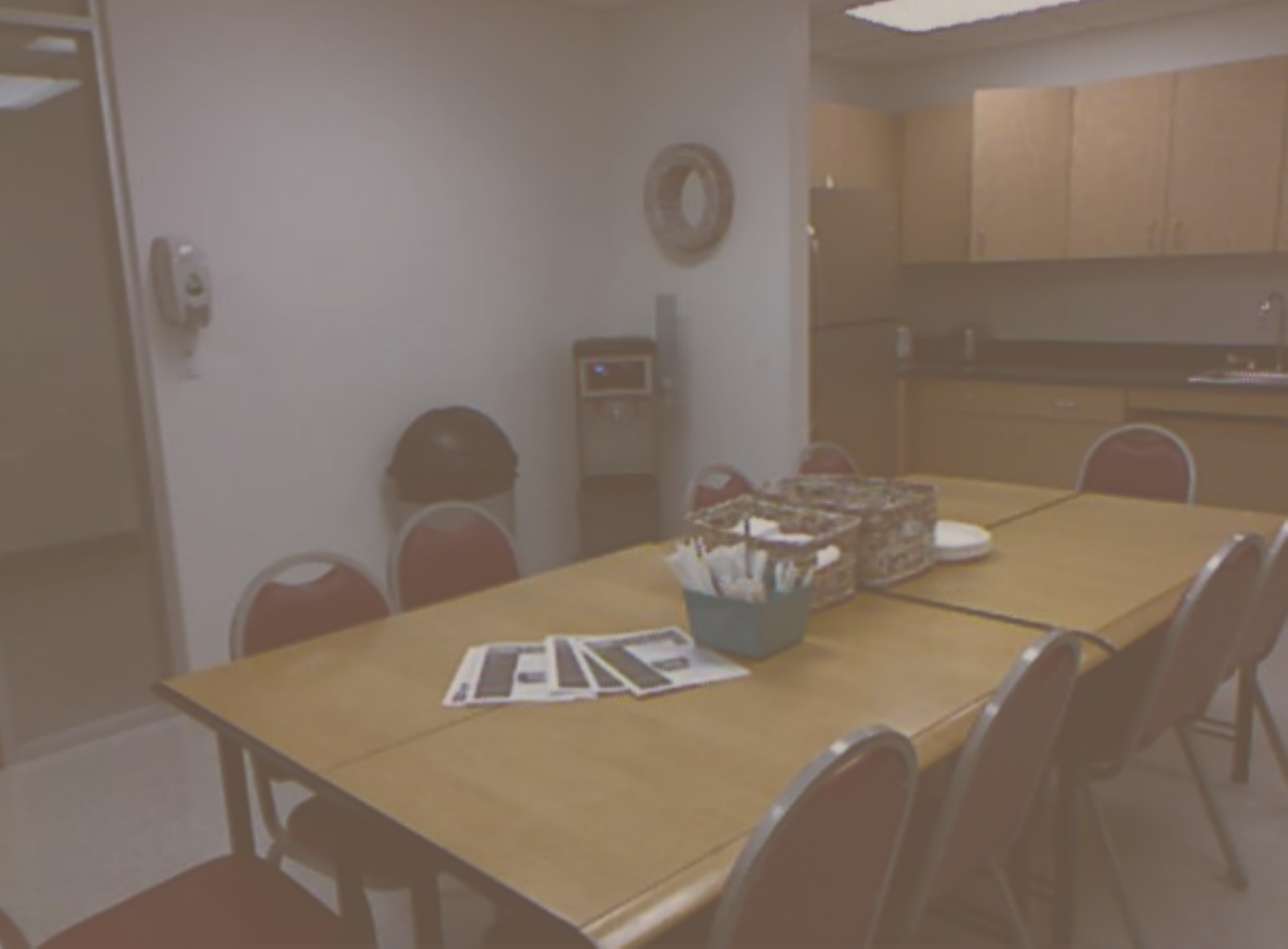}} 
\subfigure{  
\includegraphics[width=1.8cm,height=1.5cm]{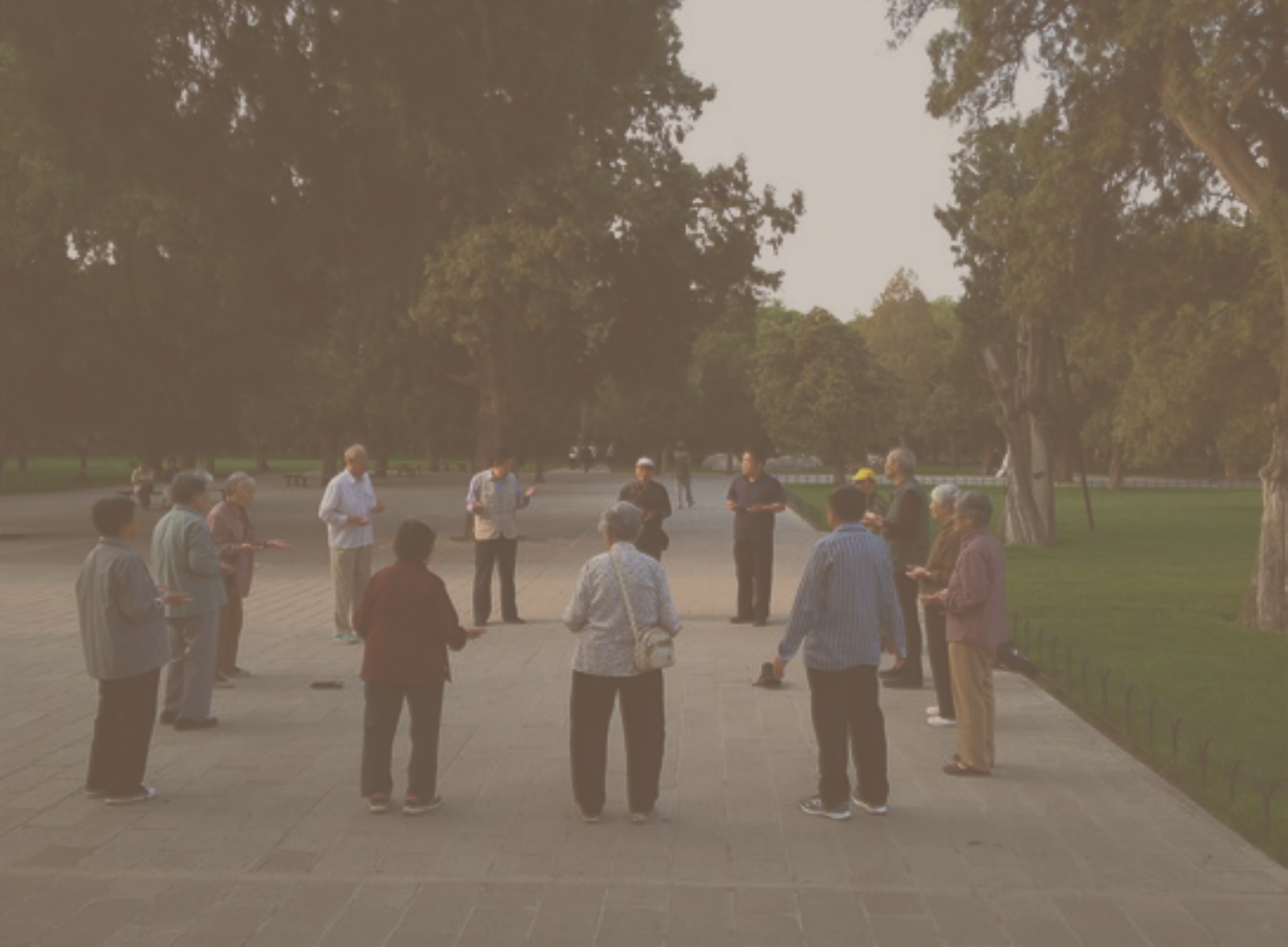}} 
\subfigure{  
\includegraphics[width=1.8cm,height=1.5cm]{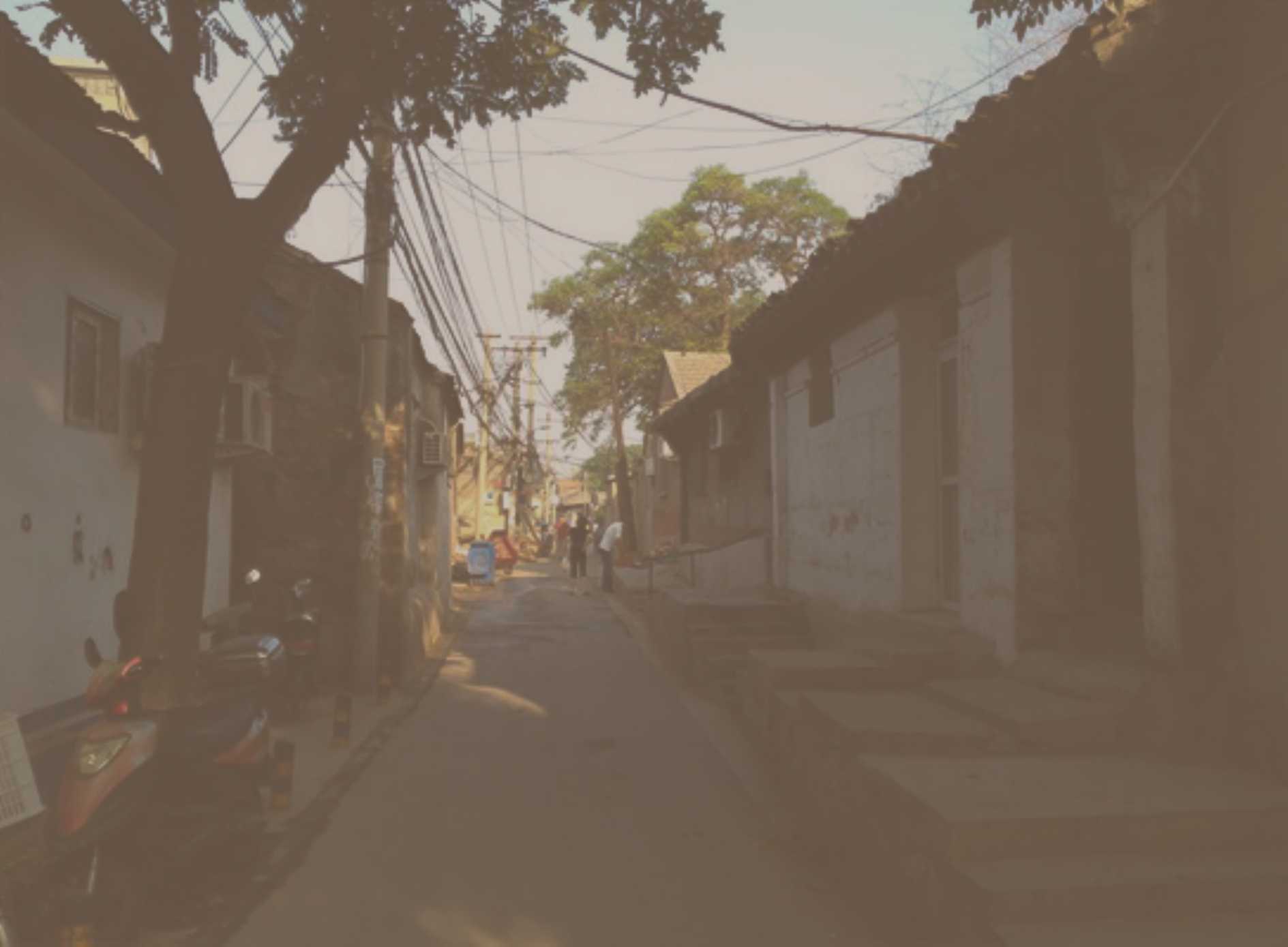}}  

\vspace{-0.2cm}

\subfigure{  
\includegraphics[width=1.8cm,height=1.5cm]{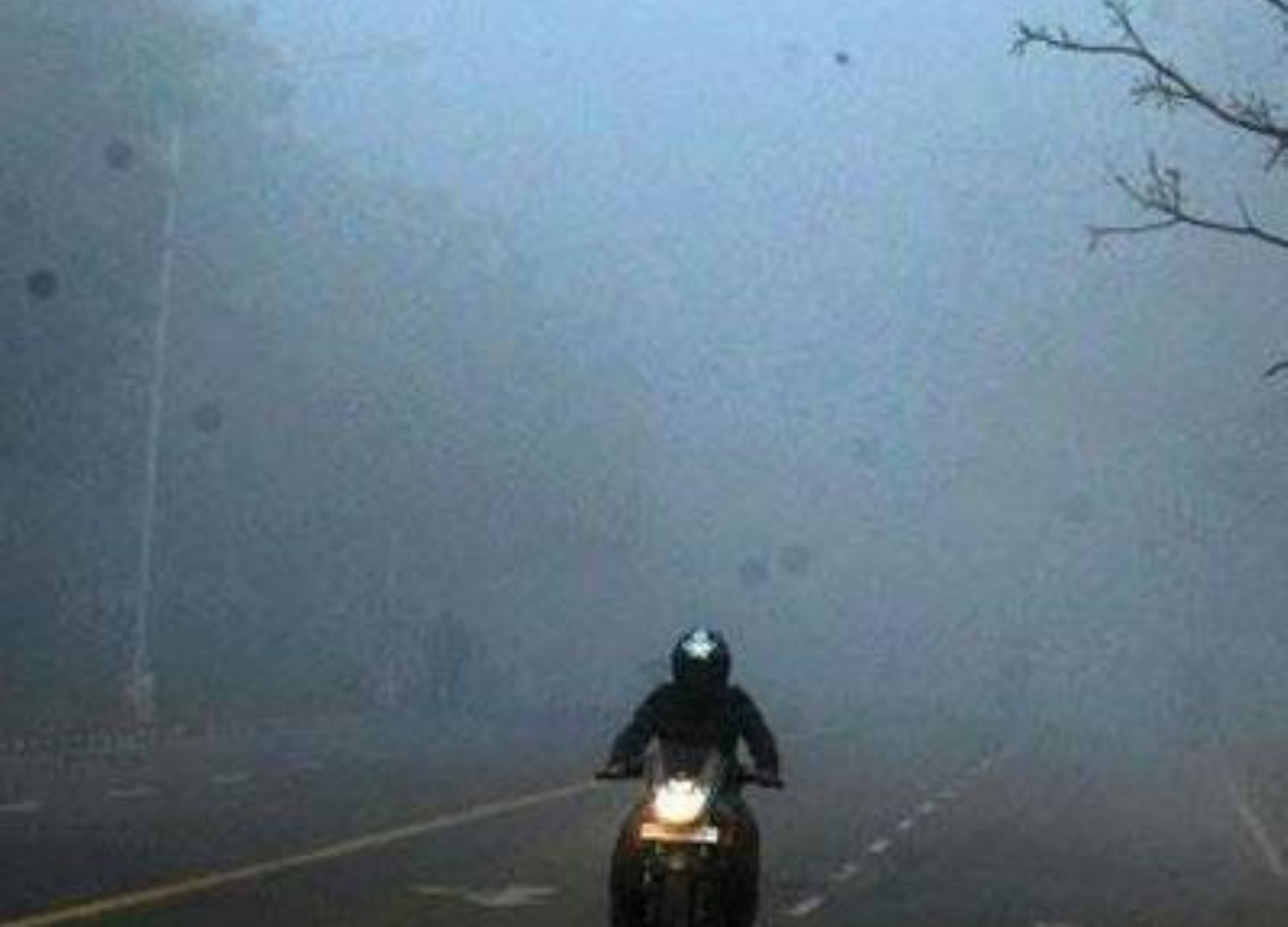}} 
\subfigure{  
\includegraphics[width=1.8cm,height=1.5cm]{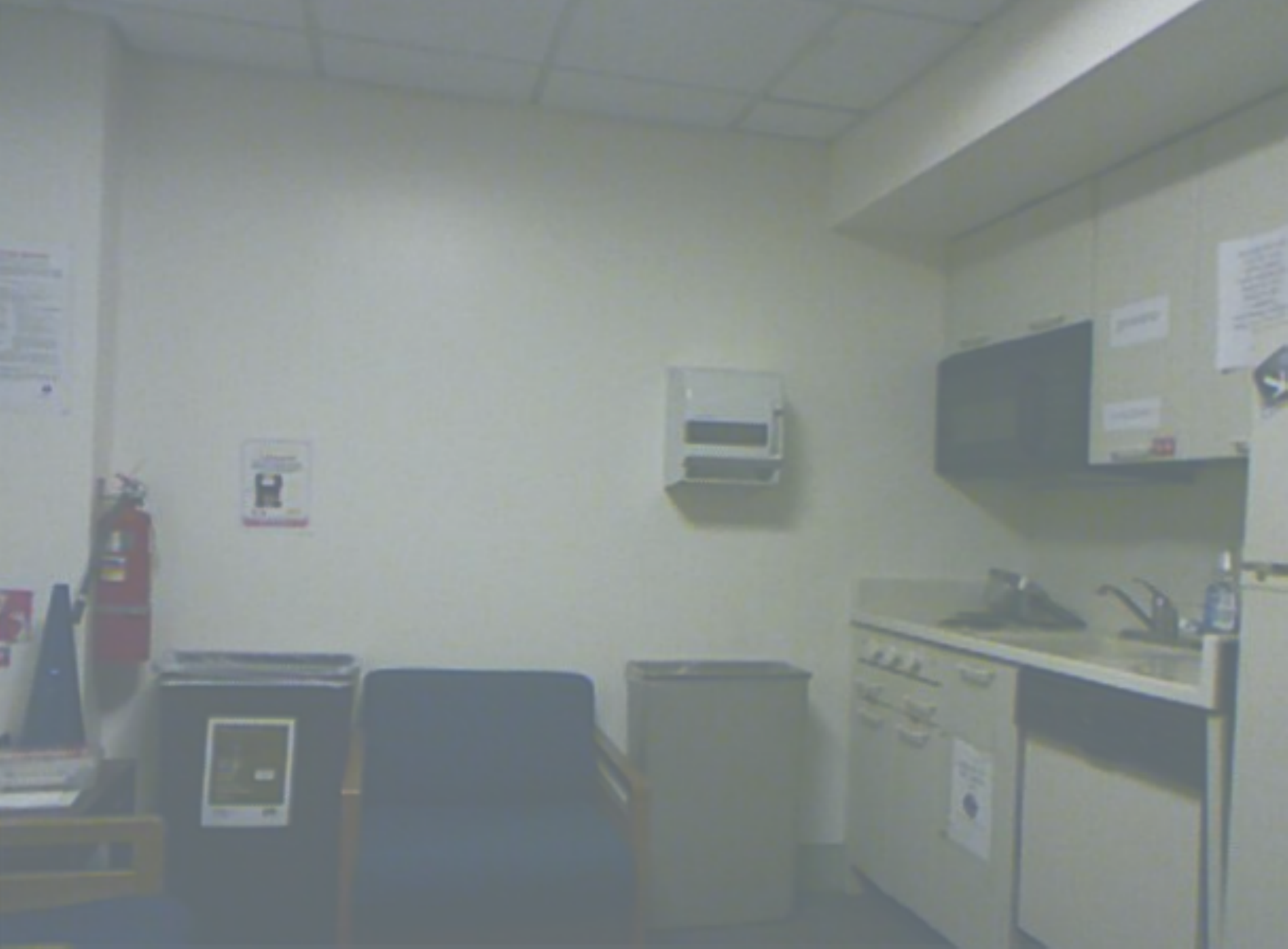}} 
\subfigure{  
\includegraphics[width=1.8cm,height=1.5cm]{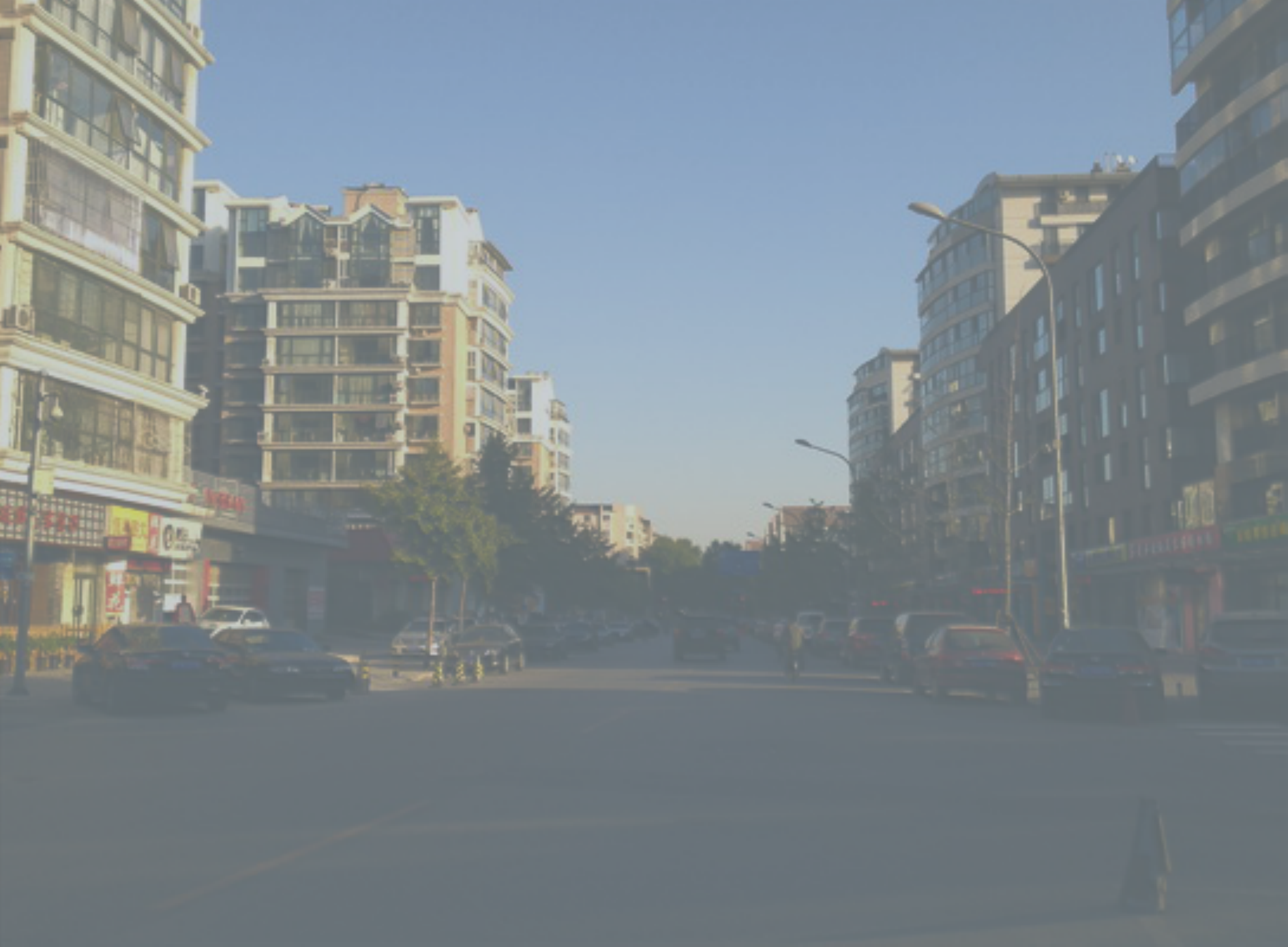}} 
\subfigure{  
\includegraphics[width=1.8cm,height=1.5cm]{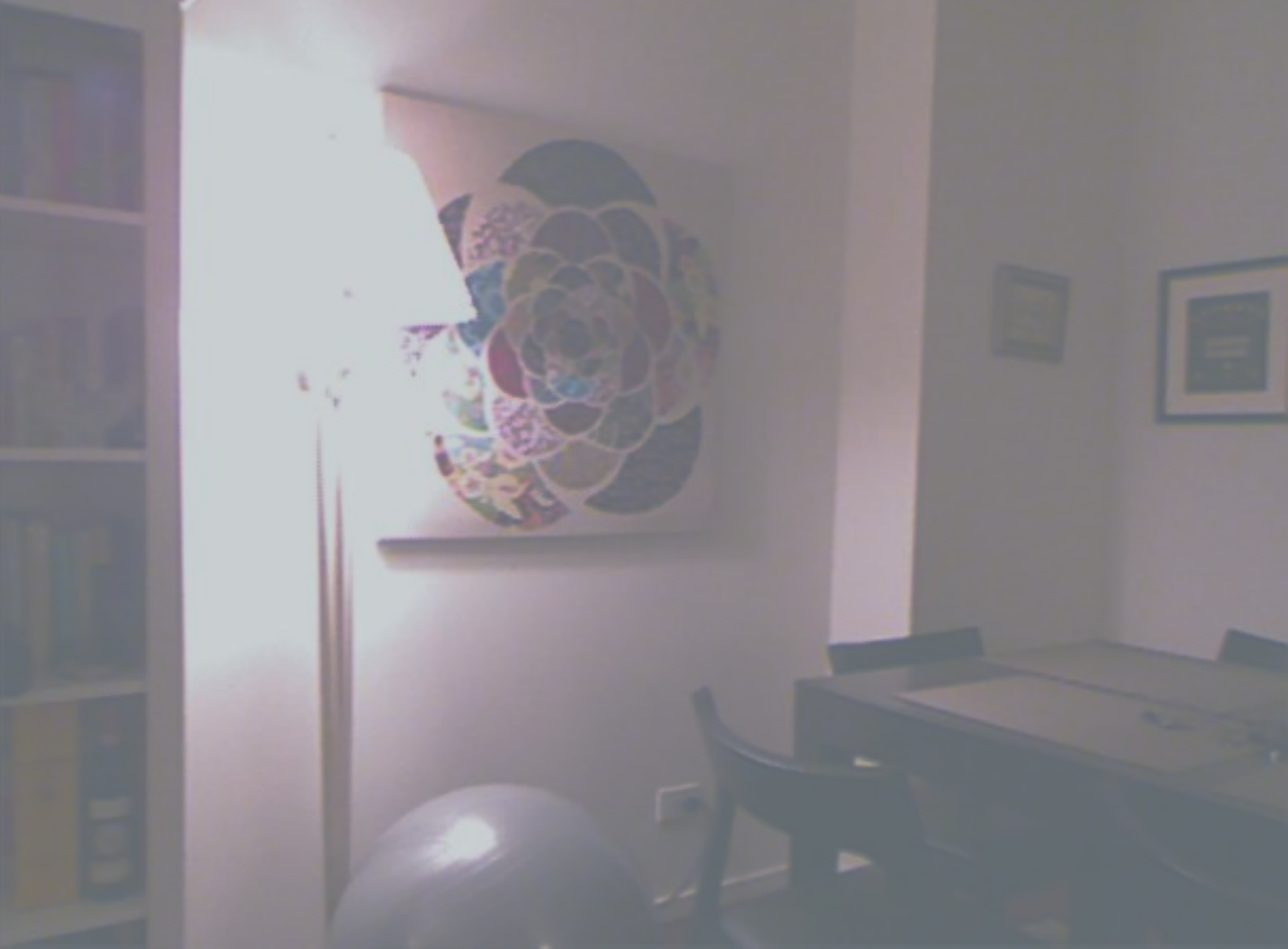}} 
\subfigure{  
\includegraphics[width=1.8cm,height=1.5cm]{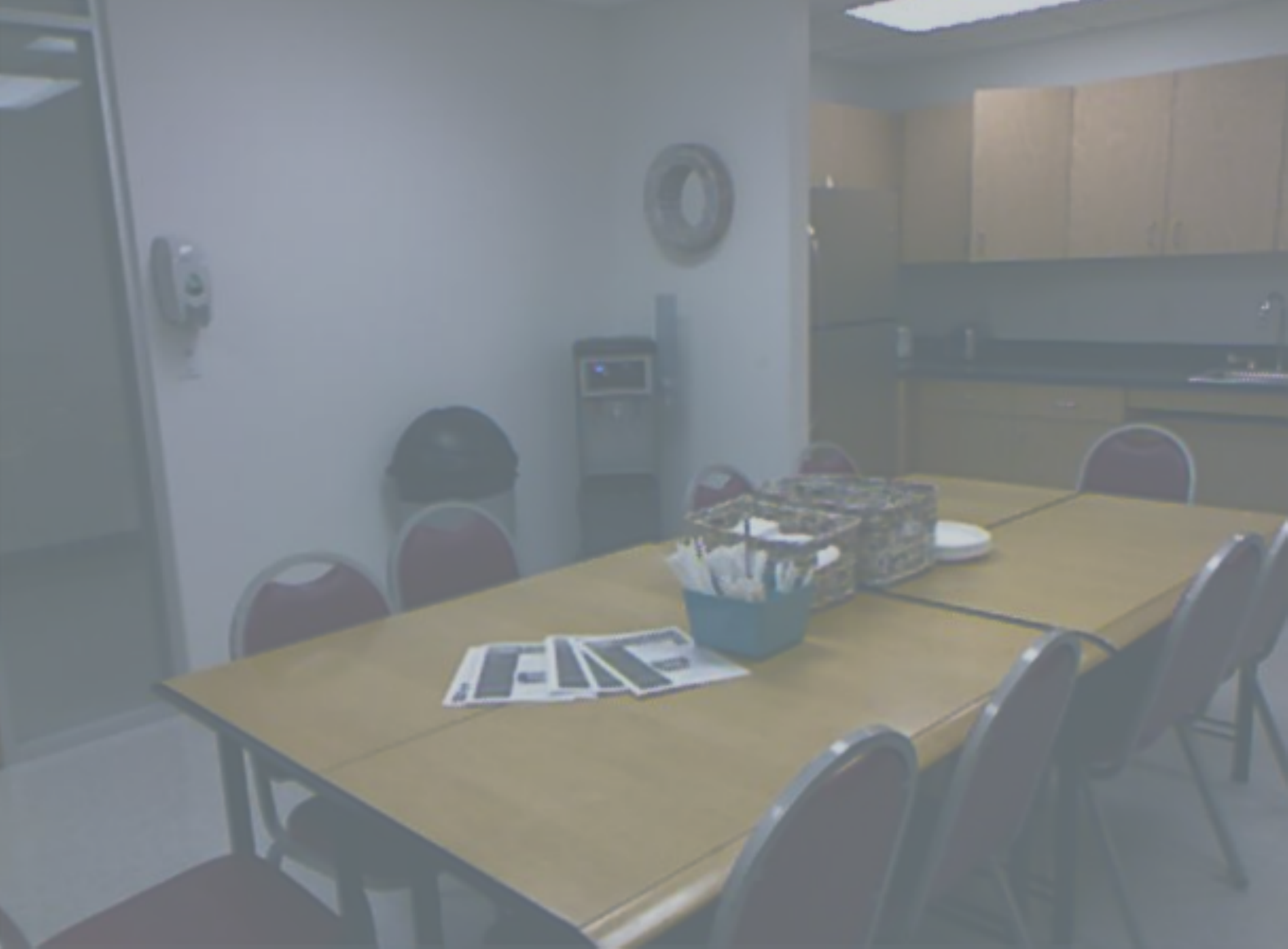}} 
\subfigure{  
\includegraphics[width=1.8cm,height=1.5cm]{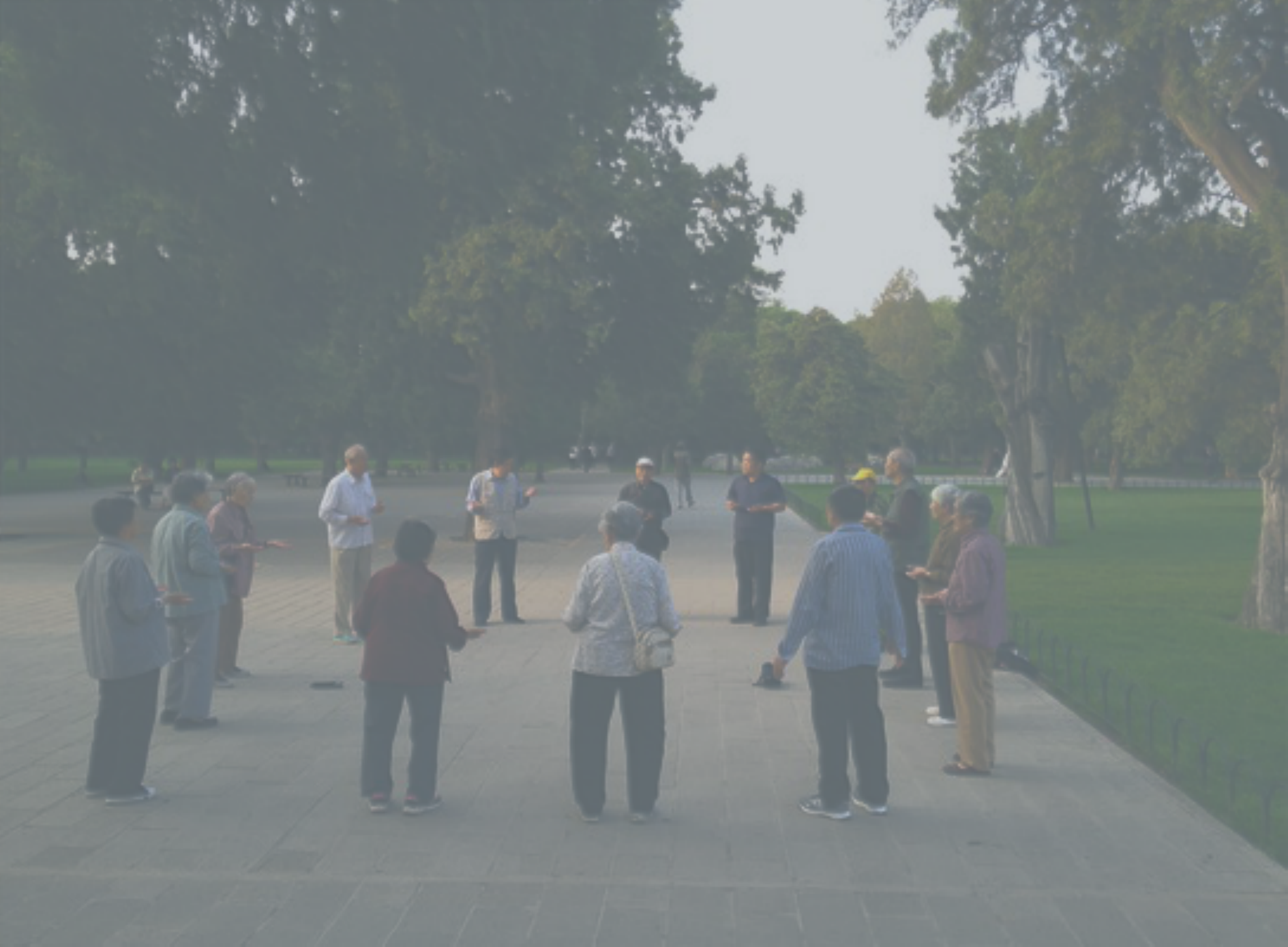}} 
\subfigure{  
\includegraphics[width=1.8cm,height=1.5cm]{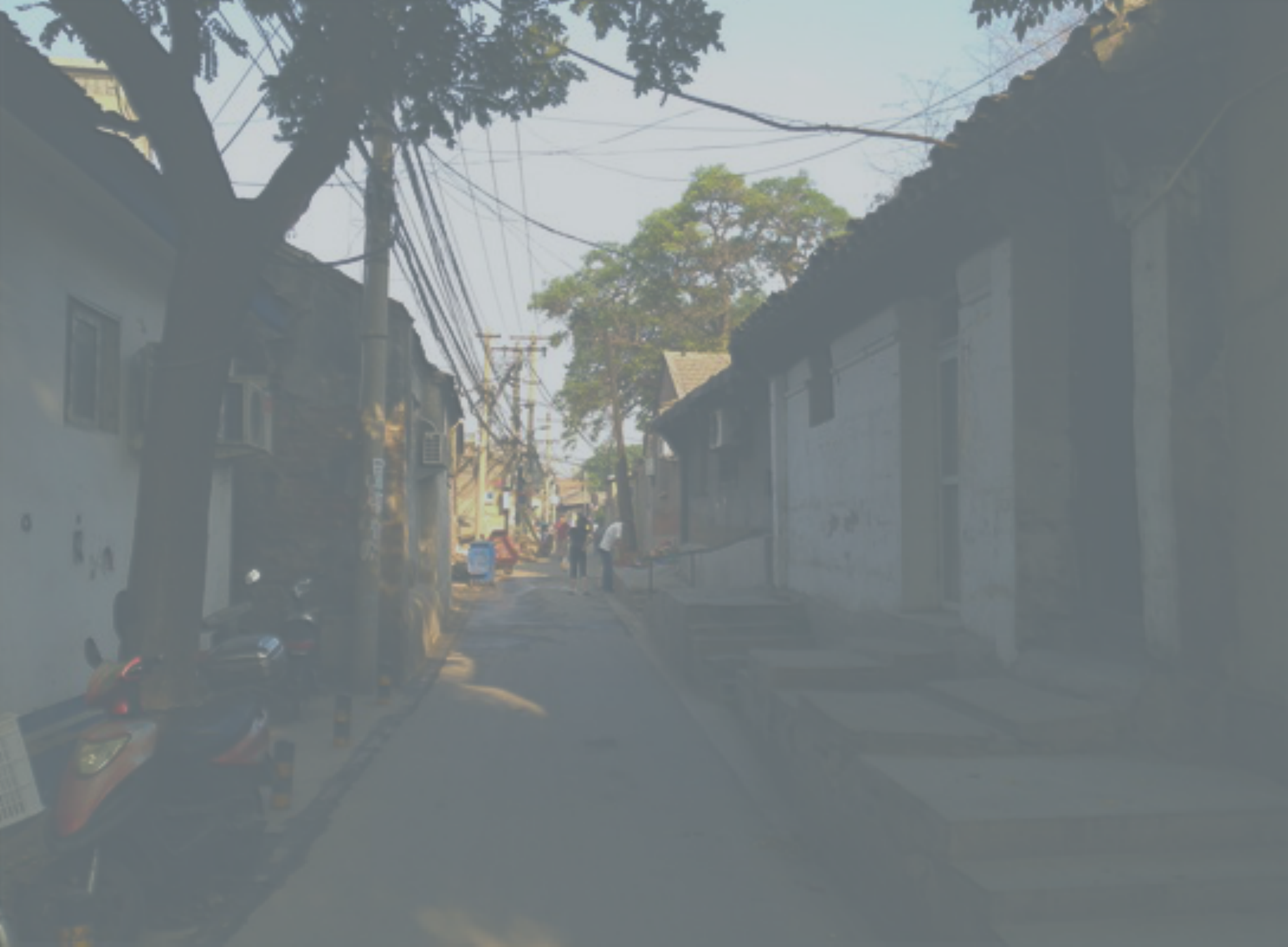}} 
	\end{center}
	\caption{\label{Figure:Airlight} Effectiveness of ATN. The first column denotes the hazy exemplar, and the reminders show the images rendered by HazeGEN.}
\end{figure*}

\subsection{Effectiveness of Airlight Transfer Network}
To verify the haze transferability of our ATN, we conduct qualitative experiments on the indoor $\mathcal{D}^{input}_{in}$ and outdoor clean scenes $\mathcal{D}^{input}_{out}$. As presented in Fig.~\ref{Figure:Airlight}, it could be seen that HazeGEN successfully transfers the airlight from the exemplar $\mathcal{D}^{e}$ to the synthetic hazy images, which demonstrates the effectiveness of ATN and the airlight consistency prior.

\begin{table*}
%\begin{small}
\centering
\caption{Ablation study on the indoor and outdoor scenes.}
\label{Table:Ablation}
%\begin{footnotesize}
\begin{tabular}{*{6}{c}}
\toprule
Scenes & Metrics & w.o. $\mathcal{L}_{S}$ & w.o. $\mathcal{L}_{A}$ & w.o. $\mathcal{L}_{adv}$& Ours \\ 
\midrule
\multicolumn{1}{c}{\multirow{2}{*}{Indoor}} & FID & 200.21 & 210.05 & 206.55 & 203.34\\
& Subjective Evaluation & 14.64\% & 27.17\% & 28.41\% &  29.78\%\\ 
\midrule
\multicolumn{1}{c}{\multirow{2}{*}{Outdoor}} & FID & 220.98 & 238.31 & 236.83 & 234.48\\
& Subjective Evaluation & 12.25\% & 14.85\% & 32.90\% & 40.00\% \\ 
\bottomrule
\end{tabular}
\end{table*}

\subsection{Ablation Study}
To elaborate the effectiveness of our loss, we conduct quantitive and subjective comparisons on both indoor and outdoor clean images by removing one of $\lambda_S$, $\lambda_A$, and $\lambda_{adv}$. The experimental setting keeps unchanged with the experiments on the reality analysis. From Table~\ref{Table:Ablation}, one could see that: 1) HazeGEN benefits from the airlight consistency prior and the adversarial loss, of which each will remarkably enhance the reality of HazeGEN; 2) removing the structure similarity prior will lead to a lower FID score and a poor visual quality. The possible reason may be that TEN cannot capture the structure information due to the removal of the structure similarity prior. In other words, the haze will be averagely added into the clean image, thus decreasing the visual quality of the images.

\section{Conclusion}
In this article, we proposed a novel knowledge-driven neural network for image hazing, called HazeGEN. The method leverages the structure prior and airlight consistency prior to generating hazy images in a learnable, unsupervised, and controllable manner. Extensive experimental results show that the promising performance of HazeGEN in both qualitative and quantitative comparisons. In the future, we plan to further extend our method in video hazing and exploit novel neural rendering methods. 

\section*{Broader Impact}

Hazy image rendering is a typical rendering task which could be applied to a wide range of applications including but not limited to gaming, filming, and image dehazing. Our work could render hazy scenes for many real-world applications and save a lot of labors, which might reduce the expertise required and labor cost in the dataset collection and game/film making. Besides the above benefits, we should care about the potential negative impacts of our method, especially, endangering the environment and lowering the employment rate. To be specific, to generate a large hazy dataset, some energies will be consumed, while causing several CO2 emissions. Moreover, it will reduce the requirement in the labors for data collection and film/game production.

%\input{render.bbl}
%\bibliographystyle{plain}
%\bibliography{render.bib}

%%%%%%%%%%%%%%%%%%%%%%%%%%%%%%%%%%%%%%%%%%%%%%%%%%%%%%%%%%%%

\end{document}